\documentclass[sigconf]{acmart}

\usepackage{amsmath}
\usepackage{algorithm}
\usepackage{algorithmic}
\usepackage{tikz}
\usetikzlibrary{shapes, shadows}
\usepackage{colortbl}
\usepackage{balance}

\setcopyright{none}
\newcommand{\set}[1]{#1}

\newcommand{\rv}[1]{\mathbf{#1}}
\newcommand{\prob}{\mathrm{Pr}}
\newcommand{\loss}{\mathit{l}}
\newcommand{\Loss}{\mathrm{L}}
\newcommand{\Gain}{\mathrm{G}}
\newcommand{\expected}[2]{\underset{#2}{\mathbb{E}}\Large[#1\Large]}

\newcommand{\DadvIN}{\set{D}^{\mathsf{A}}}
\newcommand{\DadvOUT}{\set{D'}^{\mathsf{A}}}

\newcommand{\paragraphb}[1]{\vspace{0.03in} \noindent{\bf #1} }

\graphicspath{{./images/}}

\begin{document}

\title{Machine Learning with Membership Privacy\\ using Adversarial Regularization}
\author{Milad Nasr$^1$, Reza Shokri$^2$, Amir Houmansadr$^1$\\
	$^1$ University of Massachusetts Amherst, $^2$ National University of Singapore\\
	milad@cs.umass.edu, reza@comp.nus.edu.sg, amir@cs.umass.edu}

\begin{abstract} 
Machine learning models leak information about the datasets on which they are trained.  An adversary can build an algorithm to trace the individual members of a model's training dataset.  As a fundamental inference attack, he aims to distinguish between data points that were part of the model's training set and any other data points from the same distribution.  This is known as the tracing (and also membership inference) attack.  In this paper, we focus on such attacks against black-box models, where the adversary can only observe the output of the model, but not its parameters. This is the current setting of machine learning as a service in the Internet.  

We introduce a privacy mechanism to train machine learning models that provably achieve membership privacy: the model's predictions on its training data are indistinguishable from its predictions on other data points from the same distribution.  We design a strategic mechanism where the privacy mechanism anticipates the membership inference attacks.  The objective is to train a model such that not only does it have the minimum prediction error (high utility), but also it is the most robust model against its corresponding strongest inference attack (high privacy).  We formalize this as a {\em min-max game} optimization problem, and design an adversarial training algorithm that minimizes the classification loss of the model as well as the maximum gain of the membership inference attack against it.  This strategy, which guarantees membership privacy (as prediction indistinguishability), acts also as a strong regularizer and significantly generalizes the model. 

We evaluate our privacy mechanism on deep neural networks using different benchmark datasets.  We show that our min-max strategy can mitigate the risk of membership inference attacks (close to the random guess) with a negligible cost in terms of the classification error. 
\end{abstract}

\keywords{Data privacy; Machine learning; Inference attacks; Membership privacy; Indistinguishability; Min-max game; Adversarial process}

\maketitle


\section{Introduction}

Large available datasets and powerful computing infrastructures, as well as advances in training complex machine learning models, have dramatically increased the adoption of machine learning in software systems.  Many algorithms, applications, and services that used to be built based on expert knowledge, now can be designed using much less engineering effort by relying on advanced machine learning algorithms.  Machine learning itself has also been provided as a service, to facilitate the use of this technology by system designers and application developers.  Data holders can train models using machine learning as a service (MLaaS) platforms (by Google, Amazon, Microsoft, ...) and share them with others or use them in their own applications.  The models are accessible through prediction APIs, which allow simple integration of machine learning algorithms into applications in the Internet.

A wide range of sensitive data, such as online and offline profiles of users, location traces, personal photos, speech samples, medical and clinical records, and financial portfolios, is used as input for training machine learning models.  The confidentiality and privacy of such data is of utmost importance to data holders.  Even if the training platform is trusted (e.g., using confidential computing, or by simply training the model on the data owner's servers) the remaining concern is if a model's computations (i.e., its predictions) can be exploited to endanger privacy of its sensitive training data. 

The data required for training accurate models presents serious privacy issues.  The leakage through complex machine learning models maybe less obvious, compared to, for example, linear statistics~\cite{dwork2015robust}.  However, machine learning models, similar to other types of computations, could significantly leak information about the datasets on which they are computed.  In particular, an adversary, with even black-box access to a model, can perform a membership inference~\cite{homer2008resolving, sankararaman2009genomic} (also known as the tracing~\cite{dwork2017exposed}) attack against the model to determine whether or not a target data record is a member of its training set~\cite{shokri2017membership}.  The adversary exploits the distinctive statistical features of the model's predictions on its training data.  This is a fundamental threat to data privacy, and is shown to be effective against various machine learning models and services~\cite{shokri2017membership}.

In this paper, we focus on protecting machine learning models against this exact threat: black-box membership inference attacks.  There are two major groups of existing defense mechanisms.  The first group includes simple mitigation techniques, such as limiting the model's predictions to top-k classes, therefore reducing the precision of predictions, or  regularizing the model (e.g., using L2-norm regularizers)~\cite{shokri2017membership, fredrikson2015model}.  These techniques may impose a negligible utility loss to the model.  However, they cannot guarantee any rigorous notion of privacy.  The second major group of defenses use differential privacy mechanisms~\cite{chaudhuri2011differentially, abadi2016deep, papernot2016semi, bassily2014private, papernot2018scalable}.  These mechanisms do guarantee (membership) privacy up to their privacy parameter $\epsilon$.  However, the existing mechanisms may impose a significant classification accuracy loss for protecting large models on high dimensional data for small values of $\epsilon$.  This comes from not explicitly including utility into the design objective of the privacy mechanism.  Also, it is because the differential privacy mechanisms are designed so as to guarantee input indistinguishability for {\em all} possible input training datasets (that differ in a constant number of records), and for {\em all} possible parameters/outputs of the models.  Whereas, we explicitly include utility in the objective of the privacy mechanism which protects the very existing training dataset.

\paragraph*{\bf\em Contributions} 

In this paper, we design a rigorous privacy mechanism for protecting a given training dataset, against a particular adversarial objective.  We want to train machine learning models that guarantee {\bf membership privacy}: No adversary can distinguish between the predictions of the model on its training set from the model's prediction on other data samples from the same underlying distribution, up to the privacy parameter.  This is a more targeted privacy notion than differential privacy, as we aim at a very specific (prediction) indistinguishability guarantee.  Our objective is that the privacy-preserving model should achieve membership privacy with the minimum classification loss.  

We formalize membership inference attacks and define the defender's objective for achieving membership privacy for classification models.  Based on these definitions, we design an optimization problem to minimize the classification error of the model {\em and} the inference accuracy of the strongest attack who adaptively maximizes his gain.  Therefore, this problem optimizes a composition of two conflicting objectives.  We model this optimization as a {\bf min-max privacy game} between the defense mechanism and the inference attack, similar to privacy games in other settings~\cite{hsu2013differential, manshaei2013game, shokri2012protecting, alvim2017information, jia2018attriguard}.  The solution is a model which not only is accurate but also has the maximum membership privacy against its corresponding strongest inference attack.  The adversary cannot design a better inference attack than what is already anticipated by the defender; therefore, membership privacy is guaranteed.  There does not also exist any model that, for the same level of membership privacy, can give a better accuracy.  So, maximum utility (for the same level of privacy) is also guaranteed. 

To find the solution to our optimization problem, we train the model in an {\bf adversarial process}.  The classification model maps features of a data record to classes, and computes the probability that it belongs to any class.  The primary objective of this model is to minimize prediction error.  The inference model maps a target data record, and the output of the classifier on it, to its membership probability.  The objective of the inference model is to maximize its membership inference accuracy.  To protect data privacy, we add the gain of the inference attack as a {\em regularizer} for the classifier.  Using a regularization parameter, we can control the trade-off between membership privacy and classification error.  We train the models in a similar way as generative adversarial networks~\cite{goodfellow2014generative} and other adversarial processes for machine learning~\cite{kozinski2017adversarial, miyato2015distributional, miyato2017virtual, dai2017good, odena2016semi, dumoulin2016adversarially}.  Our training algorithm can converge to an equilibrium point where the best membership inference attack against it is {\em random guess}, and this is achieved with minimum classification accuracy loss. 

We present the experimental results on deep neural networks using benchmark ML datasets as well as the datasets used in the ML privacy literature.  We compute various statistics of models' predictions on their training and test sets, in order to illustrate the worst case and the average case gaps between these statistics (which cause the privacy risk).  The gaps are reduced by several orders of magnitude when the model is trained using our min-max privacy mechanism, compared to non-privacy-preserving models. 

Our results verify our theoretical analysis that we impose only a {\bf negligible loss in classification accuracy for a significant gain in membership privacy}.  For the CIFAR100 dataset trained with Alexnet and Densenet architectures, the cost is respectively 1.1\% and 3\% drop in the prediction accuracy, relative to the regular non-privacy-preserving models.  For the Purchase100 and Texas100 datasets (used in~\cite{shokri2017membership}), the cost of membership privacy in terms of classification accuracy drop is 3.6\% and 4.4\%, respectively, for reducing the inference accuracy from 67.6\% to 51.6\% and from 63\% to 51\%, respectively.  Note that the membership privacy is maximum when the membership inference accuracy is 50\% (random guess).  

We also show that {\bf our mechanism strongly regularizes the models}, by significantly closing the gap between their training and testing accuracy, and preventing overfitting.  This directly follows from the indistinguishability of our privacy-preserving model's prediction distributions on training versus test data.  For example, on the Purchase100 dataset, we can obtain 76.5\% testing accuracy for 51.8\% membership inference accuracy.  In contrast, a standard L2-norm regularizer may provide a similar level of privacy (against the same attack) but with a 32.1\% classification accuracy.

\section{Machine Learning}


\begin{table*}[t!]
	\begin{tabular}{| p{3cm} p{11cm} |}
		\hline & \\
		Classification model &
		$f: \set{X} \longrightarrow \set{Y}$ \\[15pt]
		Loss & $\Loss(f) = \expected{\loss(f(x),y)}{(x,y) \sim \prob(\rv{X}, \rv{Y})} = \int\limits_{\set{X} \times \set{Y}} \loss(f(x),y) \, \prob(\rv{X}, \rv{Y}) \, dx \, dy$\\[15pt]
		%
		%
		Empirical loss & $\Loss_{\set{D}}(f) = \frac{1}{|\set{D}|}\sum\limits_{(x, y) \in \set{D}} \loss(f(x), y)$ \\[15pt]
		Optimization problem & $\min\limits_f \, \Loss_{\set{D}}(f) + \lambda \, R(f)$\\[15pt]
		\hline
	\end{tabular}
	\caption{Definition, loss, and optimization problem for the {\em classification model} $f$, where $x \in \set{X}$ is a data point, $y \in \set{Y}$ is a classification vector, $\set{D}$ is the model's training set, $l()$ is a loss function, $R()$ is a regularizer, and $\lambda$ is the regularization factor.\\[-15pt]}\label{tab:classification}
\end{table*}


\begin{table*}[t!]
	\begin{tabular}{| p{3cm} p{13cm} |}
		\hline & \\
		Inference model & $h : \set{X} \times \set{Y}^2 \longrightarrow [0, 1]$\\[15pt]
		Gain & $\Gain_f(h) = \expected{log(h(x, y, f(x)))}{(x,y) \sim \prob_\set{D}(\rv{X}, \rv{Y})} + \expected{\log(1 - h(x, y, f(x)))}{(x,y) \sim \prob_{\setminus\set{D}}(\rv{X}, \rv{Y})}$\\[15pt]
		Empirical gain & $\Gain_{f, \DadvIN, \DadvOUT}(h) = \frac{1}{2 |\DadvIN|} \sum_{(x,y) \in \DadvIN} \log(h(x, y, f(x))) + \frac{1}{2 |\DadvOUT|} \sum_{(x',y') \in \DadvOUT} \log( 1 - h(x', y', f(x')))$\\[15pt]
		Optimization problem & $\max\limits_h \, \Gain_{f, \DadvIN, \DadvOUT}(h)$\\[15pt]
		\hline
	\end{tabular}
	\caption{Definition, gain, and optimization problem for the {\em membership inference attack} $h$, where $f$ is the target classifier, $\prob_\set{D}(\rv{X}, \rv{Y})$ and $\prob_{\setminus\set{D}}(\rv{X}, \rv{Y})$ are the conditional probability distributions of data points in the target training set $\set{D}$ and outside it, respectively. The adversary's background knowledge is composed of datasets $\DadvIN$ (a subset of the training set $\set{D}$) and $\DadvOUT$ (samples drawn from $\prob(\rv{X}, \rv{Y})$ which are outside $\set{D}$).  See Figure~\ref{fig:attack} for the illustration of the relation between $h$ and $f$. 
	}\label{tab:attack}
\end{table*}

In this paper, we focus on training classification models using supervised learning.  Table~\ref{tab:classification} summarizes the notations and formally states the objective function of the classifier.  Let $\set{X}$ be the set of all possible data points in a $d$-dimensional space, where each dimension represents one attribute of a data point (and will be used as the input features in the classification model).  We assume there is a predefined set of $k$ classes for data points in $\set{X}$.  The objective is to find the relation between each data point and the classes as a classification function $f: \set{X} \longrightarrow \set{Y}$.  The output reflects how $f$ classifies each input into different classes.  Each element of an output $y \in \set{Y}$ is a score vector that shows the relative association of any input to different classes.  All elements of a vector $y$ are in range $[0,1]$, and are normalized such that they sum up to $1$, so they are interpreted as the probabilities that the input belongs to different classes. 

Let $\prob(\rv{X}, \rv{Y})$ represent the underlying probability distribution of all data points in the universe $\set{X} \times \set{Y}$, where $\rv{X}$ and $\rv{Y}$ are random variables for the features and the classes of data points, respectively.  The objective of a machine learning algorithm is to find a classification model $f$ that accurately represents this distribution and maps each point in $\set{X}$ to its correct class in $\set{Y}$.  We assume we have a lower-bounded real-valued loss function $\loss(f(x), y)$ that, for each data point $(x, y)$, measures the difference between $y$ and the model's prediction $f(x)$.  The machine learning objective is to find a function $f$ that minimizes the expected {\em loss}:
\begin{align}\label{eq:closs}
	\Loss(f) &= \expected{\loss(f(x),y)}{(x,y) \sim \prob(\rv{X}, \rv{Y})} 
\end{align}

We can estimate the probability function $\prob(\rv{X}, \rv{Y})$ using samples drawn from it.  These samples construct the training set $\set{D} \subset \set{X} \times \set{Y}$.  Instead of minimizing \eqref{eq:closs}, machine learning algorithms minimize the expected {\em empirical loss} of the model over its training set $\set{D}$.
\begin{align}\label{eq:closs_emp}
	\Loss_{\set{D}}(f) = \frac{1}{|\set{D}|}\sum\limits_{(x, y) \in \set{D}} \loss(f(x), y)
\end{align}

We can now state the optimization problem of learning a classification model as the following:
\begin{align}\label{eq:coptimization}
	\min\limits_f \, \Loss_{\set{D}}(f) + \lambda \, R(f)
\end{align}
where $R(f)$ is a {\em regularization} function.  

The function $R(f)$ is designed to prevent the model from overfitting to its training dataset~\cite{bishop2006pattern}.  For example, the regularization loss (penalty) increases as the parameters of the function $f$ grow arbitrarily large or co-adapt themselves to fit the particular dataset $\set{D}$ while minimizing $\Loss_{\set{D}}(f)$.  If a model overfits, it obtains a small loss on its training data, but fails to achieve a similar loss value on other data points.  By avoiding overfitting, models can generalize better to all data samples drawn from $\prob(\rv{X}, \rv{Y})$.  The regularization factor $\lambda$ controls the balance between the classification loss function and the regularizer.

For solving the optimization problem \eqref{eq:coptimization}, especially for non-convex loss functions for complex models such as deep neural networks, the commonly used method is the stochastic gradient descent algorithm~\cite{zhang2004solving, avriel2003nonlinear}.  This is an iterative algorithm where in each epoch of training, it selects a small subset (mini-batch) of the training data and updates the model (parameters) towards reducing the loss over the mini-batch.  After many epochs of training, the algorithm converges to a local minimum of the loss function.

\section{Membership Inference Attack}

The objective of membership inference attacks, also referred to as tracing attacks, is to determine whether or not a target data record is in a dataset, assuming that the attacker can observe a function over the dataset (e.g., aggregate statistics, model).

The membership inference attacks have  mostly been studied for analyzing data privacy with respect to simple statistical linear functions~\cite{homer2008resolving, sankararaman2009genomic, backes2016membership, dwork2015robust, dwork2017exposed, pyrgelis2017knock}.  The attacker compares the released statistics from the dataset, and the same statistics computed on random samples from the population, to see which one is closer to the target data record.  Alternatively, the adversary can compare the target data record and samples from the population, to see which one is closer to the released statistics.  In either case, if the target is closer to the released statistics, then there is a high chance that it was a member of the dataset.  The problem could be formulated as a hypothesis test, and the adversary can make use of likelihood ratio test to run the inference attack. 

In the case of machine learning models, the membership inference attack is not as simple, especially in the black-box setting.  The adversary needs to distinguish training set members from non-members from observing the model's predictions, which are indirect nonlinear computations on the training data.  The existing inference algorithm suggests training another machine learning model, as the inference model, to find the statistical differences between predictions on members and predictions on non-members~\cite{shokri2017membership}.  In this section, we formally present this attack and the optimization problem to model the adversary's objective.  Table~\ref{tab:attack} summarizes the notations and the optimization problem.  Figure~\ref{fig:attack} illustrates the relation between different components of a membership inference attack against machine learning models in the black-box setting. 

\tikzstyle{element} = [draw, minimum height=1.5em, minimum width=10em, rounded corners]

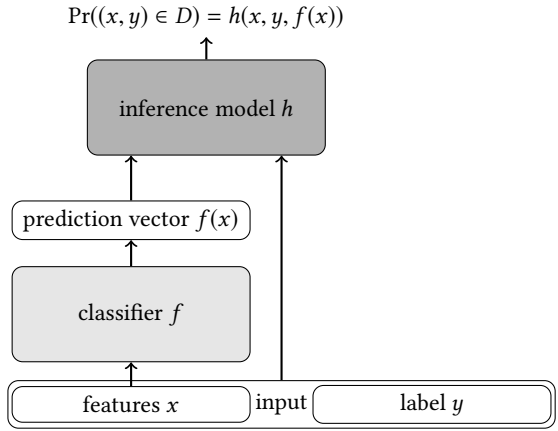
\begin{figure}[t!]
\begin{tikzpicture}
    \node (features) [element] {features $x$};
    \path (features)+(+4,0) node (label) [element] {label $y$};
	\path (features)+(+2,0) node (input) [element, minimum width=23em, minimum height=2em] {input};
    \path (features)+(0,+1.2) node (model) [element, minimum height=4em, fill=gray!20] {classifier $f$};
    \path (model)+(0,+1.25) node (output) [element] {prediction vector $f(x)$};    
    \path (output)+(+1,+1.5) node (attack) [element, minimum height=4em, fill=gray!60] {inference model $h$};
    \path (attack)+(0,+1.2) node (inference) {$\prob((x,y) \in \set{D}) = h(x, y, f(x))$};
    \draw[thick,->] (features) -- (model);
    \draw[thick,->] (model) -- (output);
    \draw[thick,->] (output.north) -- (output|-attack.south);
    \draw[thick,->] (input.north) -- (input|-attack.south);
    \draw[thick,->] (attack) -- (inference);
\end{tikzpicture}
\caption{The relation between different elements of the black-box classification model $f$ and the inference model $h$.}\label{fig:attack}
\end{figure}

Let $h$ be the inference model $h: \set{X} \times \set{Y}^2 \longrightarrow [0, 1]$.  For any data point $(x, y)$ and the model's prediction vector $f(x)$, it outputs the probability of $(x, y)$ being a member of $D$ (the training set of $f$).  Let $\prob_\set{D}(\rv{X}, \rv{Y})$ and $\prob_{\setminus\set{D}}(\rv{X}, \rv{Y})$ be the conditional probabilities of $(\rv{X}, \rv{Y})$ for samples in $\set{D}$ and outside $\set{D}$, respectively.  In an ideal setting (of knowing these conditional probability distributions), the gain function for the membership inference attack can be computed as the following.
\begin{align}\label{eq:gain}
	\Gain_f(h) &=
	\frac{1}{2} \expected{\log(h(x, y, f(x)))}{(x,y) \sim \prob_\set{D}(\rv{X}, \rv{Y})}  \nonumber \\
	&\quad + \frac{1}{2} \expected{\log(1 - h(x, y, f(x)))}{(x,y) \sim \prob_{\setminus\set{D}}(\rv{X}, \rv{Y})} \Large)
\end{align}

The two expectations compute the correctness of the inference model $h$ when the target data record is sampled from the training set, or from the rest of the universe.  In a realistic setting, the probability distribution of data points in the universe and the probability distribution over the members of the training set $\set{D}$ are not directly and accurately available to the adversary (for computing his gain).  Therefore, we compute the {\em empirical} gain of the inference model on two disjoint datasets $\DadvIN$ and $\DadvOUT$, which are sampled according to the probability distribution of the data points inside the training set and outside it, respectively.  More concretely, the dataset $\DadvIN$ could be a subset of the target training set $\set{D}$, known to the adversary.  Given these sets, the empirical gain of the membership inference model is computed as the following. 
\begin{align}\label{eq:gain_emp}
	\Gain_{f, \DadvIN, \DadvOUT}(h) &= 
	\frac{1}{2 |\DadvIN|} \sum_{(x,y) \in \DadvIN} \log(h(x, y, f(x))) \nonumber  \\ & \quad + \frac{1}{2 |\DadvOUT|} \sum_{(x',y') \in \DadvOUT} \log( 1 - h(x', y', f(x')))
\end{align}

Thus, the optimization problem for the membership inference attack is simply maximizing this empirical gain.
\begin{align}\label{eq:aoptimization}
	\max\limits_h \, \Gain_{f, \DadvIN, \DadvOUT}(h)
\end{align}

The optimization problem needs to be solved on a given target classification model $f$.  However, it is shown that it can also be trained on some shadow models, which have the same model type, architecture, and objective function as the model $f$, and are trained on data records sampled from $\prob(\rv{X}, \rv{Y})$~\cite{shokri2017membership}.

\tikzstyle{element} = [draw, minimum height=1.5em, minimum width=8em, rounded corners]

\begin{figure}[t!]
	\begin{tikzpicture}
	\node [element, minimum height=3em, fill=gray!20] (model) {classifier $f$};	
	\path (model)+(-1, +0.35) node (fo1) [] {};
	\path (model)+(1, +0.35) node (fo2) [] {};
	\path (model)+(-1,-1.6) node (traindata) [draw,thick,rotate=90,aspect=0.3,cylinder,drop shadow,fill=white,minimum width=1cm,minimum height=0.75cm] {\rotatebox{270}{$\set{D}$}};
	\path (model)+(1,-1.6) node (refdata) [draw,thick,rotate=90,aspect=0.3,cylinder,drop shadow,fill=white,minimum width=1cm,minimum height=0.75cm] {\rotatebox{270}{$\set{D'}$}};
	\draw[thick,->] (traindata.east) -- (traindata.east|-model.south);
	\draw[thick,dotted,->] (refdata.east) -- (refdata.east|-model.south);
	\path (traindata)+(+0.3, +0.75) node (x) [fill=white] {\small $x$};
	\path (refdata)+(-0.3, +0.8) node (xp) [fill=white] {\small $x'$};
	\path (model)+(0, 2) node (attack) [element, minimum height=3em, fill=gray!60] {inference model $h$};
	\path (attack)+(-1, 0.35) node (ho1) [] {};
	\path (attack)+(1, 0.35) node (ho2) [] {};
	\path (model)+(-1.4, +1) node (fx) [fill=white] {\small $f(x)$};
	\path (model)+(+1.4, +1) node (fxp) [fill=white] {\small $f(x')$};
	\draw[thick,->] (fo1) -- (fo1|-attack.south);
	\draw[thick,dotted,->] (fo2) -- (fo2|-attack.south);	
	\path (attack)+(-1.2, +1.3) node (hx) [] {\small $h(x, y, f(x)$)};
	\path (attack)+(+1.2, +1.3) node (hxp) [] {\small $h(x', y', f(x')$)};
	\draw[thick,->] (ho1) -- (ho1|-hx.south);
	\draw[thick,dotted,->] (ho2) -- (ho2|-hxp.south);	
	\path (model)+(+3.15, 0) node (loss) [] {\small Loss: $\loss(f(x), y) + $};
	\path (model)+(+4, -0.5) node (l) [] {\small $\lambda \log(h(x, y, f(x)))$};
	\draw[double,gray,<->]	(model) -- (loss);
	\path (attack)+(+3.8, 0) node (gain) [] {\small Gain: $\frac{1}{2}\log(h(x, y, f(x))) + $};	
	\path (attack)+(+4.38, -0.5) node (g) [] {\small $\frac{1}{2} \log(1 - h(x', y', f(x')))$};	
	\draw[double,gray,<->]	(attack) -- (gain);	
	\end{tikzpicture}
	\caption{Classification loss and inference gain, on the training dataset $\set{D}$ and reference dataset $\set{D'}$, in our \underline{adversarial training}. The classification loss is computed over $\set{D}$, but, the inference gain is computed on both sets.  To simplicity the illustration, the mini-batch size is set to $1$.
	}\label{fig:game}
\end{figure}
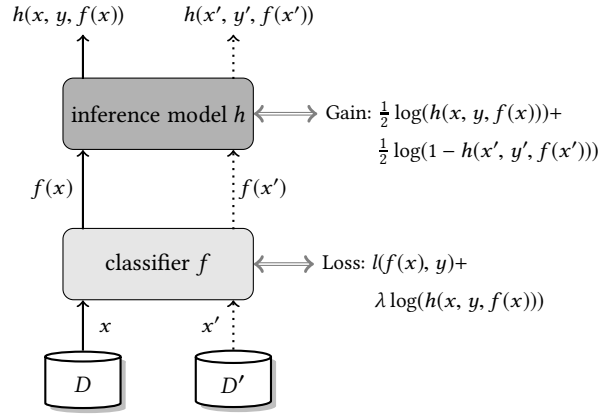
\begin{algorithm*}[t!]
	\caption{The adversarial training algorithm for machine learning with membership privacy.  This algorithm optimizes the min-max objective function \eqref{eq:game}.  Each epoch of training includes $k$ steps of the maximization part of \eqref{eq:game}, to find the best inference attack model, followed by one step of the minimization part of \eqref{eq:game} to find the best defensive classification model against such attack model.}
	\begin{algorithmic}[1]
		\FOR{number of the training epochs}
		
		\FOR{k steps}
		
		\STATE  Randomly sample a mini-batch of $m$ training data points $\{ (x_1, y_1), (x_2, y_2), \cdots, (x_m, y_m) \}$ from the training set $\set{D}$.
		
		\STATE  Randomly sample a mini-batch of $m$ reference data points $\{(x'_1, y'_1), (x'_2, y'_2), \cdots, (x'_m, y'_m)\}$ from the reference set $\set{D'}$. 
		
		\STATE Update the inference model $h$ by \underline{ascending} its stochastic gradients over its parameters $\omega$:
		\begin{align*}
		\qquad\qquad\qquad\quad \nabla_{\omega} \, \frac{\lambda}{2m}  \left( \sum_{i=1}^m \log(h(x_i, y_i, f(x_i))) + \sum_{i=1}^m \left(\log( 1 - h(x'_i, y'_i, f(x'_i))) \right) \right)
		\end{align*}
		\ENDFOR
		
		\STATE Randomly sample a fresh mini-batch of $m$ training data points  $\{ (x_1, y_1), (x_2, y_2), \cdots, (x_m, y_m) \}$ from $\set{D}$.
		
		\STATE Update the classification model $f$ by \underline{descending} its stochastic gradients over its parameters $\theta$:
		\begin{align*}
		\qquad\, \nabla_{\theta} \, \frac{1}{m} \sum_{i=1}^m \left( \loss(f(x_i), y_i)  + \lambda \, \log(h(x_i, y_i, f(x_i)))  \right)
		\end{align*}
		
		\ENDFOR
	\end{algorithmic}
	\label{alg:training}                       
\end{algorithm*}

\section{Min-Max Membership Privacy Game}
\label{sec:game}

The adversary always has the upper hand.  He adapts his inference attack to his target model in order to maximize his gain with respect to this {\em existing} classification model.  This means that a defense mechanism will be eventually broken if it is designed with respect to a particular attack, without anticipating and preparing for the (strongest) attack against itself.  The conflicting objectives of the defender and the adversary can be modeled as a privacy game~\cite{hsu2013differential, manshaei2013game, shokri2012protecting, alvim2017information}.  In our particular setting, while the adversary tries to get the maximum inference gain, the defender needs to find the classification model that not only minimizes its loss, but also minimizes the adversary's maximum gain.  This is a min-max game.  

The privacy objective of the classification model is to minimize its privacy loss with respect to the worst case (i.e., maximum inference gain) attack.  It is easy to achieve this by simply making the output of the model independent of its input, at the cost of destroying the utility of the classifier.  Thus, we update the training objective of the classification model as minimizing privacy loss with respect to the strongest inference attack, {\em with} minimum classification loss.  This results in designing the optimal privacy mechanism which is also utility maximizing.

We formalize the joint privacy and classification objectives in the following min-max optimization problem.

\begin{align}\label{eq:game}
	\underbrace{ \min\limits_f \left( \Loss_{\set{D}}(f) + \lambda \, \underbrace{ \max\limits_h \, \Gain_{f, \set{D}, \set{D}'}(h)}_{\text{optimal inference}} \right)}_{\text{optimal privacy-preserving classification}}
\end{align}

The inner maximization finds the strongest inference model $h$ against a given classification model $f$.  The outer minimization finds the strongest defensive classification model $f$ against a given $h$.  The parameter $\lambda$ controls the importance of optimizing classification accuracy versus membership privacy.  The inference attack term which is multiplied by $\lambda$ acts as a regularizer for the classification model.  In other words, it prevents the classification model to arbitrarily adapt itself to its training data at the cost of leaking information about the training data to the inference attack model.  Note that \eqref{eq:game} is equivalent to \eqref{eq:coptimization}, if we set $R(f)$ to \eqref{eq:aoptimization}.

These two optimizations need to be solved jointly to find the equilibrium point.  For arbitrarily complex functions $f$ and $h$, this game can be solved numerically using the stochastic gradient descent algorithm (similar to the case of generative adversarial networks~\cite{goodfellow2014generative}).  The training involves two datasets: the training set $\set{D}$, which will be used to train the classifier, and a disjoint reference set $\set{D'}$ that, similar to the training set, contains samples from $\prob(\rv{X}, \rv{Y})$. 

Algorithm~\ref{alg:training} presents the pseudo-code of the adversarial training of the classifier $f$ on $\set{D}$\textemdash against its best inference attack model $h$.  In each epoch of training, the two models $f$ and $h$ are alternatively trained to find their best responses against each other through solving the nested optimizations in \eqref{eq:game}.  In the inner optimization step: for a fixed classifier $f$, the inference model is trained to distinguish the predictions of $f$ on its training set $\set{D}$ from predictions of the same model $f$ on reference set $\set{D'}$.  This step maximizes the empirical inference gain $\Gain_{f, \set{D}, \set{D}'}(h)$.  In the outer optimization step: for a fixed inference attack $h$, the classifier is trained on $\set{D}$, with the adversary's gain function acting as a regularizer.  This minimizes the empirical classification loss $\Loss_{\set{D}}(f) + \lambda \, \Gain_{f, \set{D}, \set{D}'}(h)$.  We want this algorithm to converge to the equilibrium point of the min-max game that solves~\eqref{eq:game}.

\paragraph*{\bf\em Theoretical Analysis.}
Our ultimate objective is to train a classification model $f$ such that it has indistinguishably similar output distributions for data members of its training set versus the non-members.  We make use of the theoretical analysis of the generative adversarial networks~\cite{goodfellow2014generative} to reason about how Algorithm~\ref{alg:training} tries to converge to such privacy-preserving model.  For a given classification model $f$, let $p_{f}$ be the probability distribution of its output (i.e., prediction vector) on its training data $\set{D}$, and let $p'_{f}$ be the probability distribution of the output of $f$ on any data points outside the training dataset (i.e., $X\times Y \setminus D$).

For a given classifier $f$, the {\em optimal attack} model maximizes~\eqref{eq:gain}, which can be expanded to the following. 
\begin{align}
	&\Gain_f(h) = \frac{1}{2} ( \int_{x,y} \prob_\set{D}(x, y) \, p_{f}(f(x)) \log( h(x, y, f(x))) dxdy   \nonumber\\
	\quad &+ \int_{x',y'} \prob_{\setminus\set{D}}(x', y') \, p'_{f}(f(x')) \log( 1-h(x', y', f(x')))dx'dy' ) 
\end{align}

The maximum value of $\Gain_f(h)$ is achievable by the optimal inference model $h^*_f$ with enough learning capacity, and is equal to
\begin{align}
h^*_{f}(x,y,f(x))= \frac{\prob_{\set{D}}(x, y) \, p_{f}(f(x))}{\prob_{\set{D}}(x,y) \, p_{f}(f(x))+\prob_{\setminus\set{D}}(x,y) \, p'_{f}(f(x))}
\end{align}

This combines what is already known (to the adversary) about the distribution of data inside and outside the training set, and what can be learned from the predictions of the model about its training set.  Given that the training set is sampled from the underlying probability distribution $\prob(\rv{X}, \rv{Y})$, and assuming that the underlying distribution of the training data is a-priori unknown to the adversary, the optimal inference model is the following. 
\begin{align}
h^*_{f}(x,y,f(x))= \frac{p_{f}(f(x))}{p_{f}(f(x))+p'_{f}(f(x))}
\end{align}

This means that the best strategy of the adversary is to determine membership by comparing the probability that the prediction $f(x)$ comes from distribution $p_f$ or alternatively from $p'_f$. 

Given the optimal strategy of adversary against any classifier, we design the {\em optimal classifier} as the best response to the inference attack.  The privacy-preserving classification task has two objectives~\eqref{eq:game}: minimizing both the classification loss $\Loss_D(f)$ and the privacy loss $\Gain_{f,D,D'}(h^*_f)$.  In the state space of all classification models $f$ that have the same classification loss $\Loss_D(f)$, the min-max game~\eqref{eq:game} will be reduced to solving $\min_f\max_h \, \Gain_{f, \set{D}, \set{D}'}(h)$ which is then computed as:
\begin{align}
\min\limits_f \, \max\limits_h &\expected{\log(h(x, y, f(x)))}{(x,y) \sim \prob_{\set{D}}(\rv{X}, \rv{Y})} \nonumber \\
	& + \expected{\log(1 - h(x, y, f(x)))}{(x,y) \sim \prob_{\setminus\set{D}}(\rv{X}, \rv{Y}) }
\end{align}

According to Theorem~1 in~\cite{goodfellow2014generative}, the optimal function $f^*$ is the global minimization function if and only if $p_{f^*}=p'_{f^*}$.  This means that for a fixed classification loss and with enough learning capacity for model $f$, the training algorithm minimizes the privacy loss by making the two distributions $p_{f^*}$ and $p'_{f^*}$ indistinguishable.  This implies that the optimal classifier pushes the membership inference probability $h^*_{f}(x,y,f(x))$ to converge to $0.5$, i.e., random guess.  According to  Proposition~2 in ~\cite{goodfellow2014generative}, we can prove that the stochastic gradient descent algorithm of Algorithm~\ref{alg:training} eventually converges to the equilibrium of the min-max game~\eqref{eq:game}.  To summarize, the solution will be a classification model with minimum classification loss such that the strongest inference attack against it cannot distinguish its training set members from non-members by observing the model's predictions on them. 


\section{Experiments}
\label{sec:experiments}

In this section, we apply our method to several different classification tasks using various neural network structures.  We implemented our method using Pytorch\footnote{\url{http://pytorch.org/}}.  The purpose of this section is to  empirically show the robustness of our privacy-preserving model against inference attacks and its negligible classification loss. 

\subsection{Datasets}

We use three datasets: a major machine learning benchmark dataset (CIFAR100), and two datasets (Purchase100, Texas100)  which are used in the original membership inference attack against machine learning models~\cite{shokri2017membership}.

\paragraphb{CIFAR100.} 
This is a major benchmark dataset used to evaluate image recognition algorithms~\cite{krizhevsky2009learning}.  The dataset contains 60,000 images, each composed of $32\times32$ color pixels.  The records are clustered into 100 classes, where each class represents one object. 

\paragraphb{Purchase100.}
This dataset is based on  Kaggle's ``acquire valued shopper'' challenge.~\footnote{\url{https://www.kaggle.com/c/acquire-valued-shoppers-challenge/data}} The dataset includes shopping records for several thousand individuals. The goal of the challenge is to find offer discounts to attract new shoppers to buy new products.  Courtesy of the authors~\cite{shokri2017membership}, we obtained the processed and simplified version of this dataset.  Each data record corresponds to one costumer and has 600 binary features (each corresponding to one item).  Each feature reflects if the item is purchased by the costumer or not. The data is clustered into 100 classes and the task is to predict the class for each costumer. The dataset contains 197,324 data records. 

\paragraphb{Texas100.}
This dataset includes hospital discharge data. The records in the dataset contain information about inpatient stays in several health facilities published by the Texas Department of State Health Services.  Data records have features about the external causes of injury (e.g., suicide, drug misuse), the diagnosis (e.g., schizophrenia, illegal abortion), the procedures the patient
underwent (e.g., surgery), and generic information such as  gender, age, race, hospital ID, and length of stay. Courtesy of the authors~\cite{shokri2017membership}, we obtained the processed dataset, which contains 67,330 records and 6,170 binary features which represent the 100 most frequent medical procedures.  The records are clustered into 100 classes, each representing a different type of patient. 

\subsection{Classification Models}

For the CIFAR100 dataset, we used two different neural network architectures.  (1) Alexnet architecture~\cite{krizhevsky2012imagenet}, trained with Adam optimizer\cite{kingma2014adam} with learning rate 0.0001, and 100 epochs of training. (2) DenseNet architecture~\cite{huang2017densely}, trained with stochastic gradient descent (SGD) for 300 epochs, with learning rate 0.001 from epoch 0 to 100, 0.0001 from 100 to 200, and 0.00001 from 200 to 300. Following their architectures, both these models are regularized. Alexnet uses Dropout (0.2), and Densenet uses L2-norm regularization (5e-4).

For the Purchase100 dataset, we used a 4-layer fully connected neural network with layer sizes [1024, 512, 256, 100].  We used Tanh activation functions, similar to~\cite{shokri2017membership}.  We initialized all of parameters with a random normal distribution with mean 0 and standard deviation 0.01.  We trained the model for 50 epochs.

For the Texas dataset, we used a 5-layer fully connected neural network with layer sizes [2048, 1024, 512, 256, 100], with Tanh activation functions.  We initialized all of parameters with a random normal distribution with mean 0 and standard deviation 0.01.  We trained the model for 50 epochs. 

Table~\ref{tab:setting} shows the number of training data as well as reference data samples which we used in our experiments for different datasets.  It also reports the adversarial regularization factor $\lambda$ which was used in our experiments. 

\tikzstyle{element} = [draw, rounded corners, fill=gray]
\begin{figure}[t!]
	\begin{tikzpicture}
	\node [] (fx) {$f(x)$};	
	\path (fx)+(+4, 0) node (y) [] {$y$};
	
	\path (fx)+(0, 0.75) node (fl1) [element, minimum width=5em] {};
	\path (fl1)+(0, 0.25) node (t) [] {$100\times1024$};	
	\path (fl1)+(0, 0.5) node (fl2) [element, fill=gray, minimum width=15em] {};
	\path (fl2)+(0, 0.25) node (t) [] {$1024\times512$};	
	\path (fl2)+(0, 0.5) node (fl3) [element, fill=gray, minimum width=10em] {};
	\path (fl3)+(0, 0.25) node (t) [] {$512\times64$};	
	\path (fl3)+(2, 0.5) node (t) [draw, dashed, rounded corners, minimum width=15em] {};
	
	\path (fl3)+(0, 0.5) node (fl4) [element, fill=gray, minimum width=3em] {};
	
	\path (y)+(0, 1.25) node (yl1) [element, fill=gray, minimum width=5em] {};
	\path (yl1)+(0, 0.25) node (t) [] {$100\times512$};	
	\path (yl1)+(0, 0.5) node (yl2) [element, fill=gray, minimum width=10em] {};
	\path (yl2)+(0, 0.25) node (t) [] {$512\times64$};	
	\path (yl2)+(0, 0.5) node (yl3) [element, fill=gray, minimum width=3em] {};
	
	\path (y)+(-2, 2.75) node (h1) [element, fill=gray, minimum width=8em] {};
	\path (h1)+(0, 0.25) node (t) [] {$256\times64$};	
	\path (h1)+(0, -0.25) node (t) [] {$128\times256$};	
	\path (h1)+(0, 0.5) node (h2) [element, fill=gray, minimum width=3em] {};
	\path (h2)+(0, 0.25) node (t) [] {$64\times1$};	
	\path (h2)+(0, 0.5) node (h3) [element, fill=gray] {};
	\path (h3)+(0, 0.75) node (out) [] {$h(x,y,f(x))$};
	
	\draw[thick,->]	(fx) -- (fl1);
	\draw[thick,->]	(y) -- (yl1);
	\draw[thick,->]	(h3) -- (out);
	
	\end{tikzpicture}
	\caption{The neural network architecture for the inference attack model. Each layer is fully connected to its subsequent layer. The size of each fully connected layer is provided.}\label{fig:attackmodel}
\end{figure}
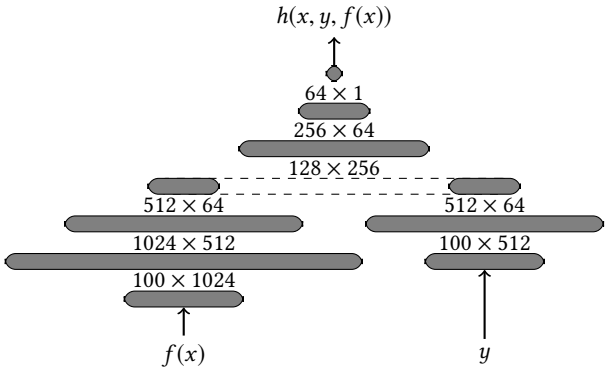

\subsection{Inference Attack Model}

For the inference model, we also make use of neural networks.  Figure~\ref{fig:attackmodel} illustrates the architecture of our inference neural network.  The objective of the attack model is to compute the membership probability of a target record $(x,y)$ in the training set of the classification model $f$.  The attack model inputs $(x,y)$ as well as the prediction vector of the classification model on it, i.e., $f(x)$.  We design the inference attack model with three  separate fully connected sub-networks.  One network of layer sizes [100,1024,512,64] operates on the prediction vector $f(x)$.  One network of layer sizes [100,512,64] operates on the label which is one-hot coded (all elements are 0 except the one that corresponds to the label index).  The third (common) network operates on the concatenation of the output of the first two networks and has layer sizes of [256,64,1].  In contrast to~\cite{shokri2017membership} which trains $k$  membership inference attack models (one per class), we design only a single model for the inference attack.  The architecture of our attack model, notably its last (common) layers, enables capturing the relation between the class and the predictions of the model for training set members versus non-members.

We use ReLu as the activation function in the network.  All weights are initialized with normal distribution with mean 0 and standard deviation 0.01, and all biases are initialized to 0.  We use Adam optimizer with learning rate 0.001. We make sure every training batch for the attack model has the same number of member and non-member instances to prevent the attack model to be biased toward one side.
  
Table~\ref{tab:setting} shows the number of known members of the training set, $\DadvIN$, and known non-member data points, $\DadvOUT$, that we assume for the adversary, which is used for training his attack model.  The larger these sets  are (especially the $\DadvIN$ set), the more knowledge is assumed for the attacker.  As we are evaluating our defense mechanism, we assume a strong adversary who knows a substantial fraction of the training set and tries to infer the membership of the rest of it.

\begin{table}[t!]
	\centering
	\begin{tabular}{|c|c|c|c|c|c|}
		\hline
		Dataset & $|D|$ &  $|D'|$ &  $\lambda$ & $|\DadvIN|$ & $|\DadvOUT|$  \\
		\hline\hline
		Purchase100 & 20,000 &  20,000 & 3 & 5,000 & 20,000 \\
		Texas100 & 10,000 &  5,000 & 2 & 5,000 & 10,000 \\
		CIFAR100 & 50,000 &  5,000 & 6 & 25,000 & 5,000 \\
		\hline
	\end{tabular}\\[2pt]
	\caption{Experimental setup, including the size of the training set $\set{D}$ and reference set $\set{D'}$ in Algorithm~\ref{alg:training}, the adversarial regularization factor $\lambda$, as well as the size of the adversary's known members $\DadvIN$ of the classifier's training set and known non-members $\DadvOUT$.}
	\label{tab:setting}
\end{table}
\begin{figure}[t]
	\centering
	\includegraphics[width=0.95\columnwidth]{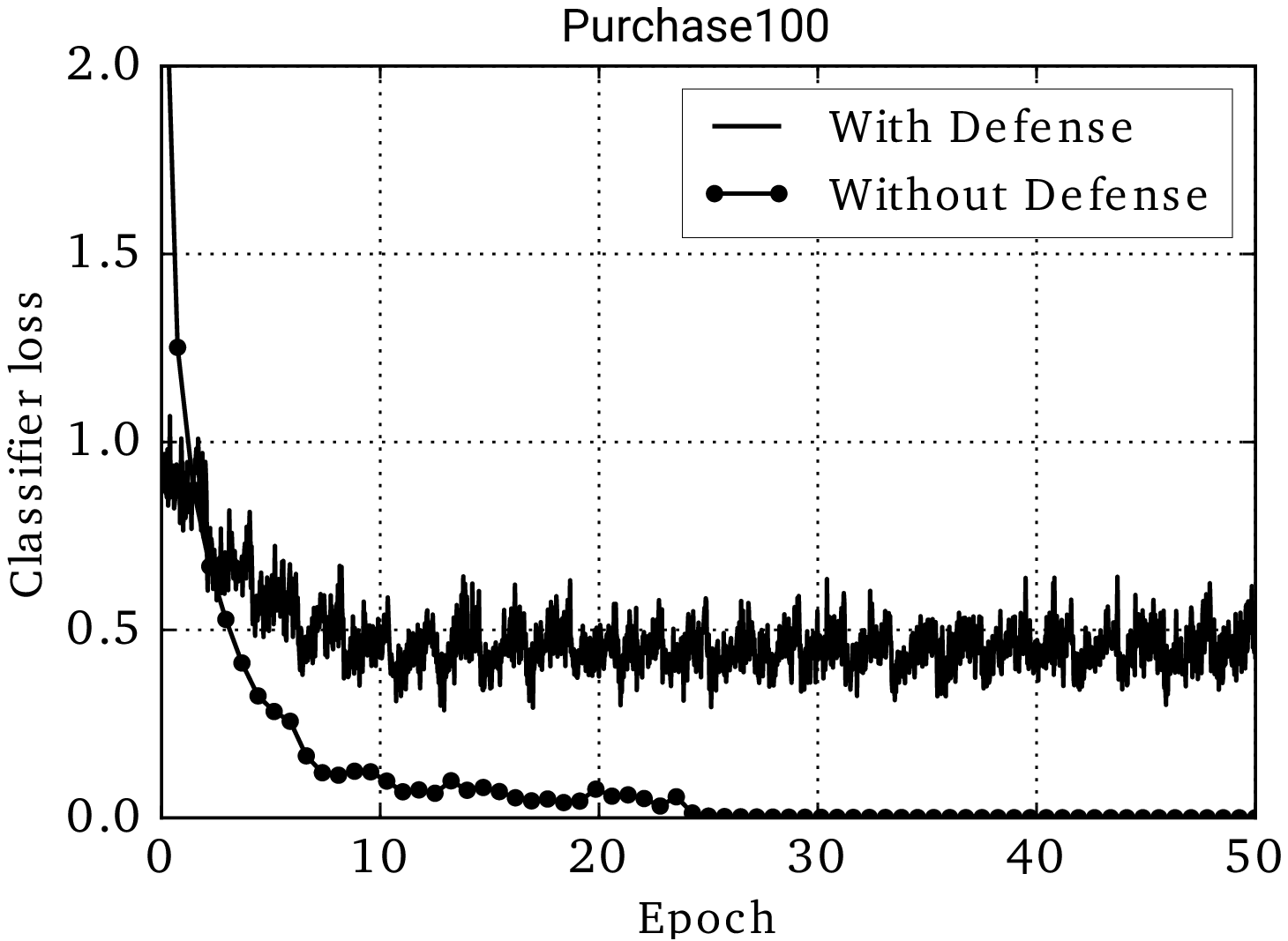}\\[1pc]
	\includegraphics[width=0.95\columnwidth]{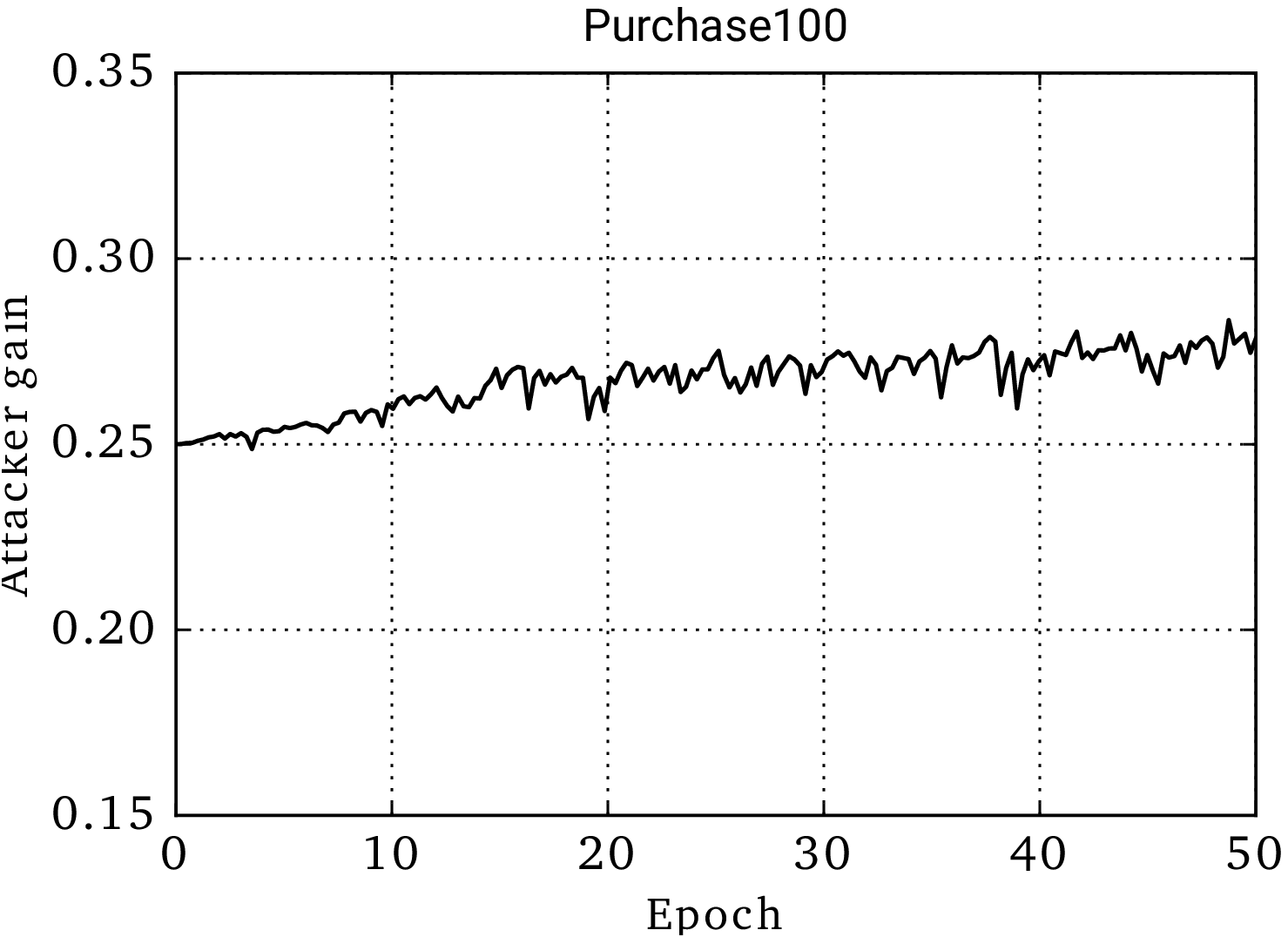}
	\caption{The trajectory of the classification loss during training with/without defense mechanism, as well as the inference attack gain, using the Purchase100 dataset.}
	\label{fig:trajectories}
\end{figure}

\begin{table*}[t]
	\begin{tabular}{|l|p{1.2cm}p{1.2cm}p{1.2cm}|p{1.2cm}p{1.2cm}p{1.2cm}|}
		\hline
		&\multicolumn{3}{|c|}{Without defense} & \multicolumn{3}{|c|}{With defense} \\
		\hline
		Dataset  & Training accuracy & Testing accuracy  & Attack  accuracy  & Training accuracy & Testing accuracy & Attack  accuracy\\
		\hline\hline
		Purchase100 &  $100\%$ & \cellcolor{green!20}$80.1\%$ & \cellcolor{red!20}$67.6\%$ & $92.2\%$ & \cellcolor{green!20}$76.5\%$ & \cellcolor{red!20}$51.6\%$ \\ 
		Texas100 &  $81.6\%$ & \cellcolor{green!20}$51.9\%$ & \cellcolor{red!20}$63\%$ &  $55\%$ & \cellcolor{green!20}$47.5\%$ & \cellcolor{red!20}$51.0\%$ \\
		CIFAR100- Alexnet &  $99\%$ & \cellcolor{green!20}$44.7\%$ & \cellcolor{red!20}$53.2\%$  & $66.3\%$ & \cellcolor{green!20}$43.6\%$ & \cellcolor{red!20}$50.7\%$ \\
		CIFAR100- DenseNET &  $100\%$ & \cellcolor{green!20}$70.6\%$  & \cellcolor{red!20}$54.5\%$ & $80.3\%$ & \cellcolor{green!20}$67.6\%$ & \cellcolor{red!20}$51.0\%$ \\
		\hline
	\end{tabular}\\[2pt]
	\caption{Comparison of membership privacy and training/test accuracy of a  classification model (without defense), and a privacy-preserving model (with defense) on four different models/datasets.  Compare the two cases with respect to the \underline{trade-off between testing accuracy and attack accuracy}.  See Table~\ref{tab:setting} for the experimental setup.}    
	\label{tab:results}
\end{table*}

\begin{table}[t]
  \begin{tabular}{|c|c|c|c|}
    \hline
    $\lambda$ & Training & Testing & Attack \\
    & accuracy & accuracy & accuracy \\
    \hline\hline
    \cellcolor{gray!20}$0$ (no defense)& \cellcolor{gray!20} $100\%$ & \cellcolor{gray!20}$80.1\%$ & \cellcolor{gray!20}$67.6\%$\\
    $1$ &$98.7\%$ &  $78.3\%$ & $57.0\%$ \\
    $2$ & $96.7\%$ & $77.4\%$ & $55.0\%$\\ 
    $3$ & $92.2\%$ & $76.5\%$ & $51.8\%$\\
    $10$ & $76.3\%$ & $70.1\%$ & $50.6\%$\\
    \hline
  \end{tabular}\\[2pt]
  \caption{The effect of the adversarial regularization factor $\lambda$, used in the min-max optimization~\eqref{eq:game}, which also acts as our privacy parameter, on the defense mechanism trained on the Purchase100 dataset.}\label{tab:lambda}
\end{table}
\begin{table}
	\centering
	\begin{tabular}{|c|c|c|c|}
		\hline
		L2-regularization & Training & Testing & Attack\\
		factor & accuracy & accuracy & accuracy\\
		\hline\hline
		\cellcolor{gray!20}$0$ (no regularization) & \cellcolor{gray!20}$100\%$ & \cellcolor{gray!20}$80.1\%$ & \cellcolor{gray!20}$67.6\%$ \\		
		$0.001$ & $86\%$ & $81.3\%$ & $60\%$ \\
		$0.005$ & $74\%$ & $70.2\%$ & $56\%$ \\
		$0.01$ & $34\%$ & $32.1\%$ & $50.6\%$ \\
		\hline
	\end{tabular}\\[2pt]
	\caption{The results of using a $L2-$regularization as a mitigation technique for membership inference attack. The model is trained on the Purchase100 dataset.  Compare these results with those in Table~\ref{tab:results} which shows what we can achieve using the strategic min-max optimization.}
	\label{tab:L2regularization}
\end{table}

\begin{table}
	\begin{tabular}{|c|c|c|}
		\hline
		Reference set size & Testing accuracy & Attack accuracy \\
		\hline\hline
		1,000 & $80.0\%$& $59.2\%$ \\
		5,000 & $77.4\%$ & $52.8\%$\\
		10,000 & $76.8\%$ & $52.4\%$\\
		20,000 & $76.5\%$ & $51.6\%$ \\
		30,000 & $76.4\%$ & $50.6\%$ \\
		\hline
	\end{tabular}\\[2pt]
	\caption{The effect of the size of the reference set $\set{D'}$ on the defense mechanism for the Purchase100 dataset.  Note that (as also shown in Table~\ref{tab:setting}) the size of the training set is 20,000.}
	\label{tab:refsize}
\end{table}


\begin{figure}[th!]
	\centering
	\includegraphics[width=0.90\columnwidth]{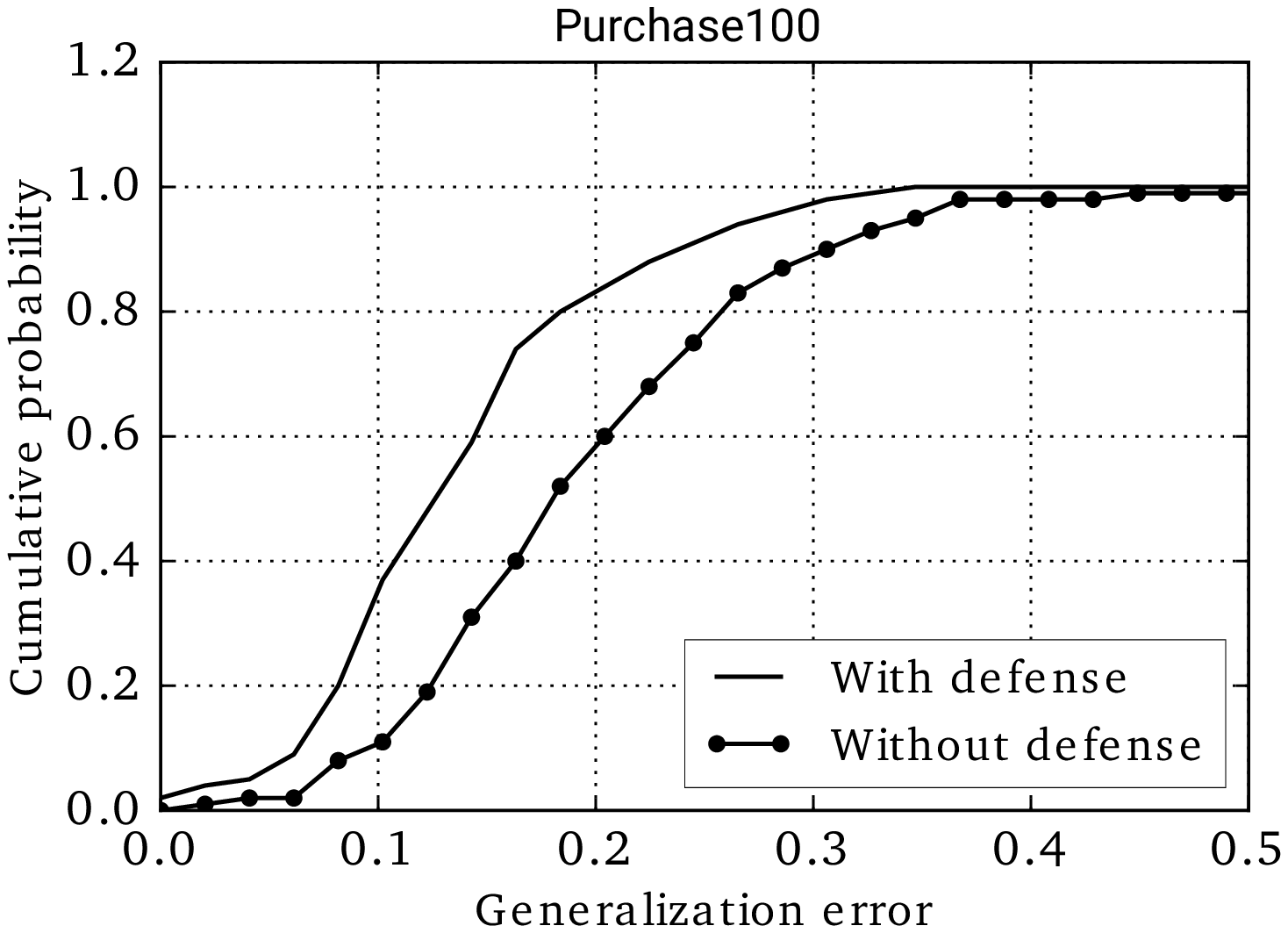}\\[15pt]
	\includegraphics[width=0.90\columnwidth]{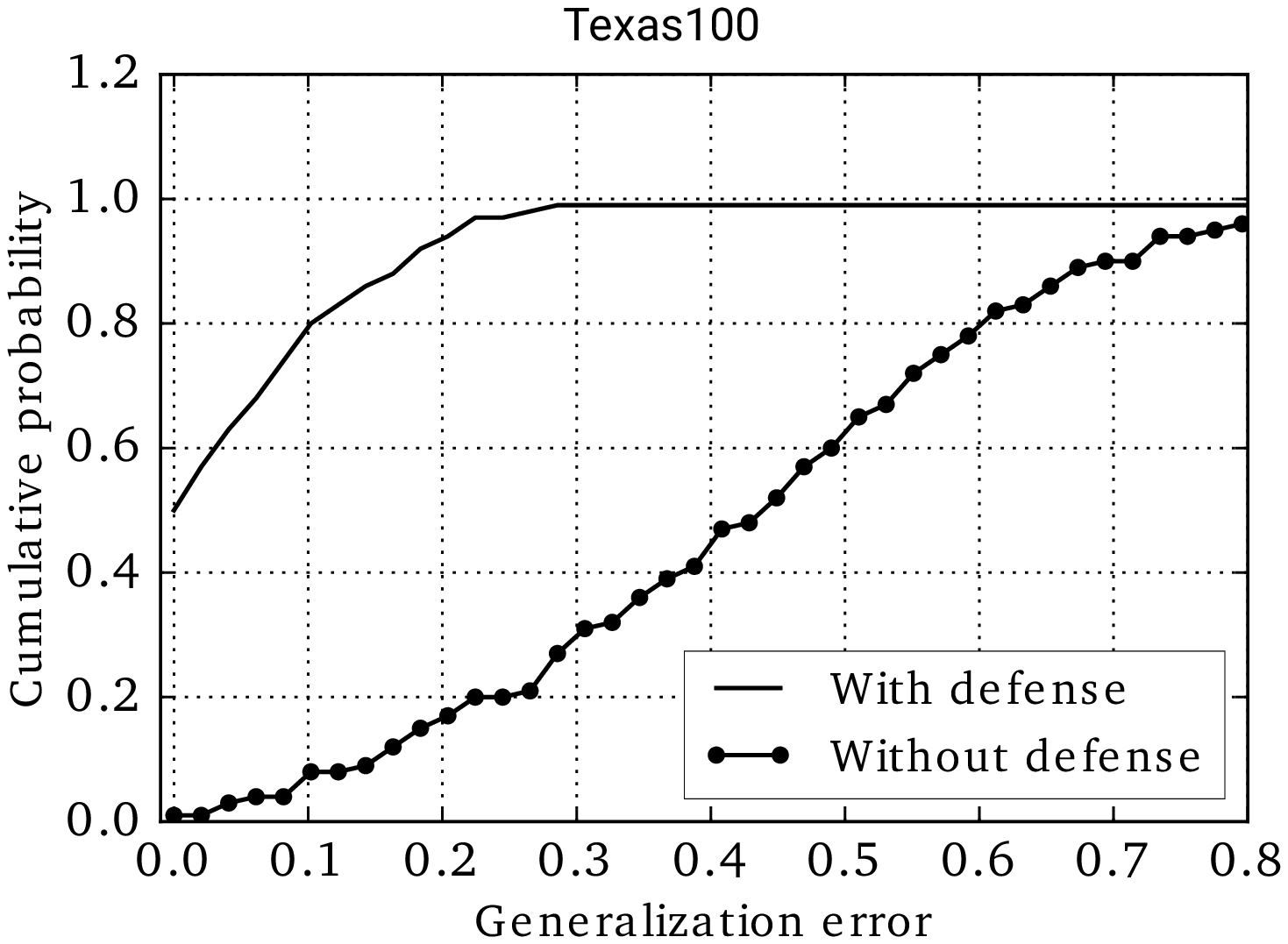}\\[15pt]
	\includegraphics[width=0.90\columnwidth]{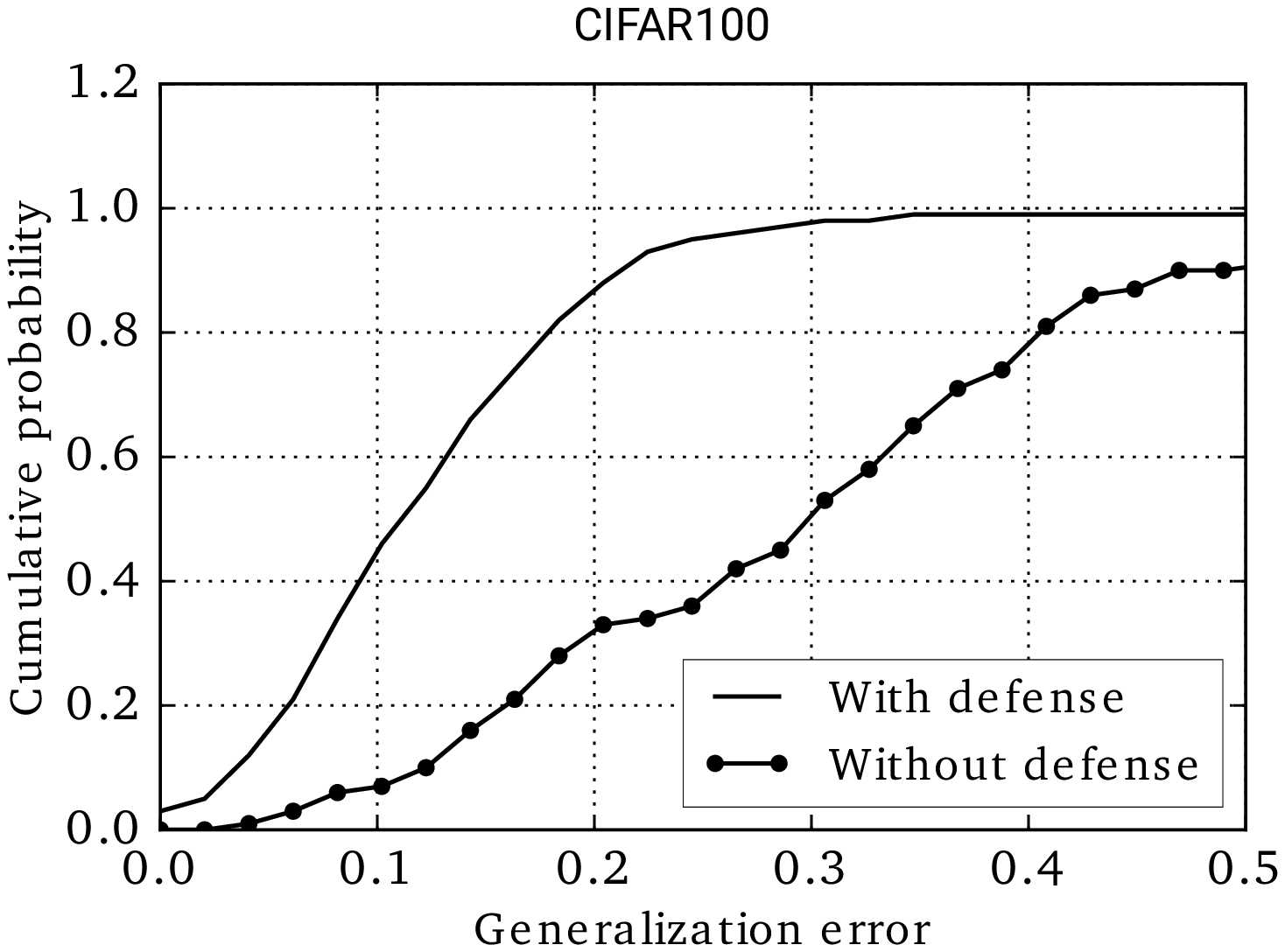}
	\caption{The empirical CDF of the \underline{generalization error} of classification models across different classes, for regular models (without defense) versus privacy-preserving models (with defense).  We compute generalization error as the difference between the training and testing accuracy of the model~\cite{hardt2015train}.  The y-axis is the fraction of classes that have generalization error less than x-axis.  The curves that lean towards left have a smaller generalization error. \\[15pt]}
	\label{fig:genralizationerror}
\end{figure}


\begin{figure}[th!]
	\centering
    \includegraphics[width=0.90\columnwidth]{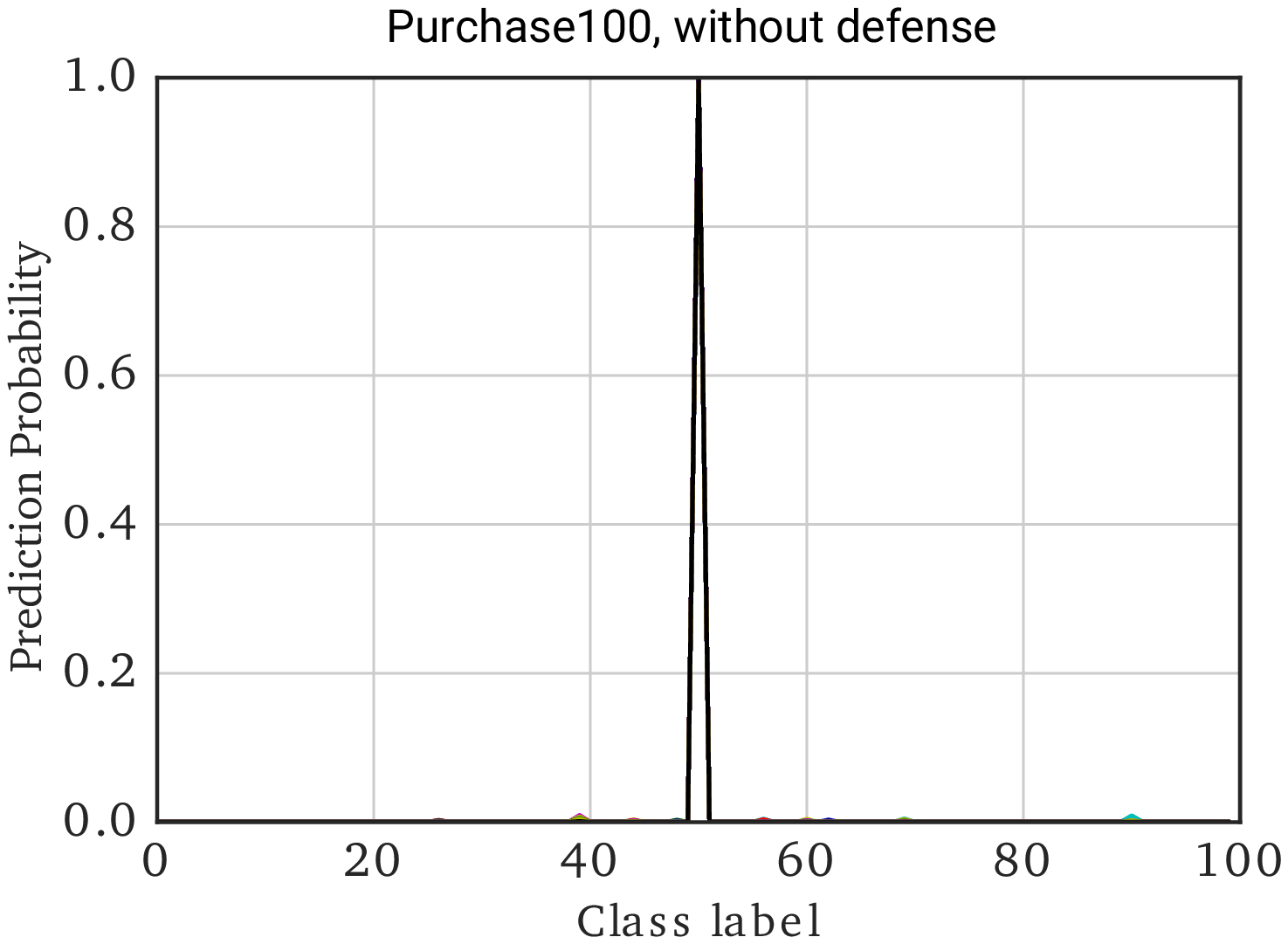}\\[15pt]
    \includegraphics[width=0.90\columnwidth]{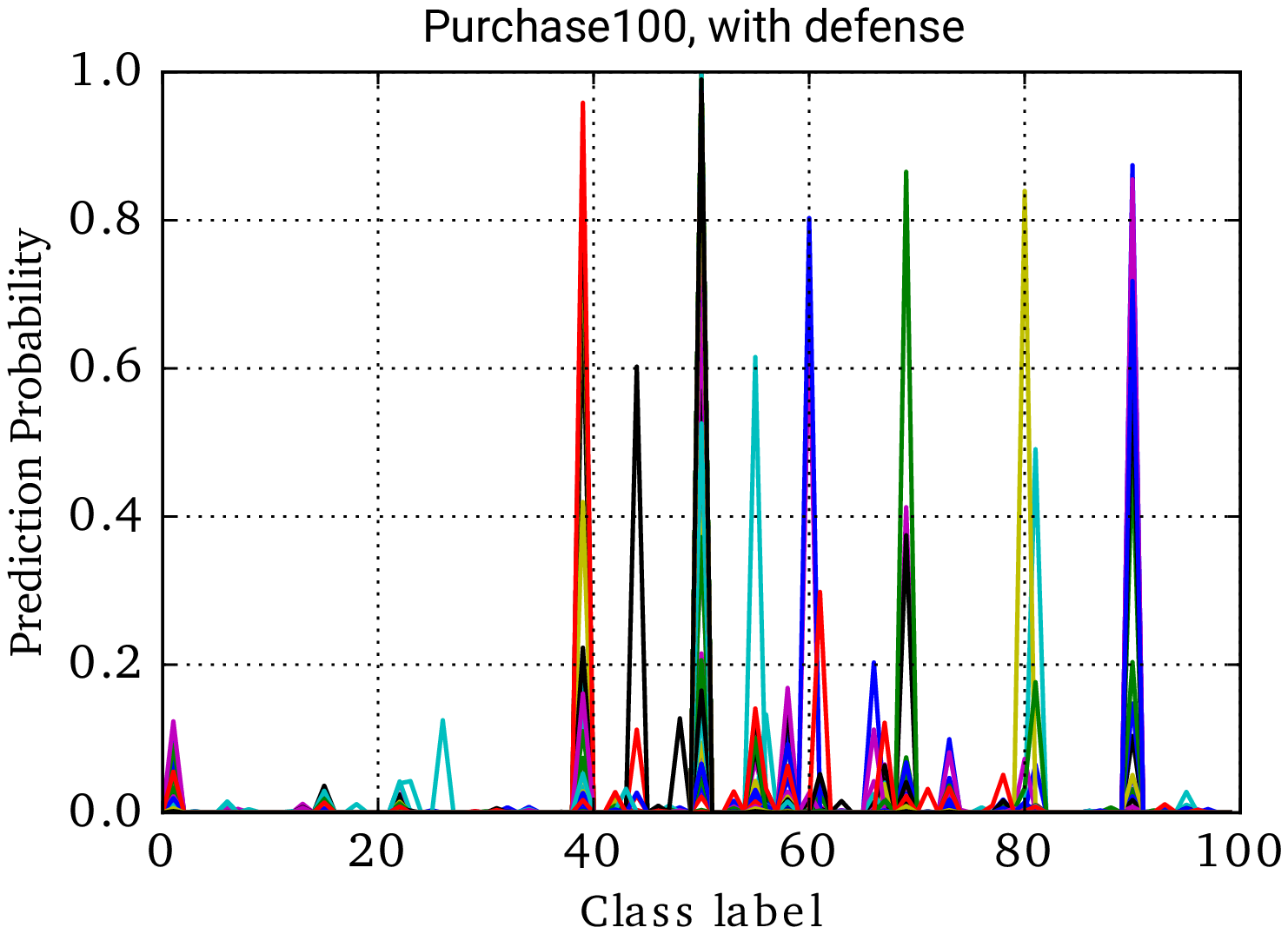}\\[15pt]
    \includegraphics[width=0.90\columnwidth]{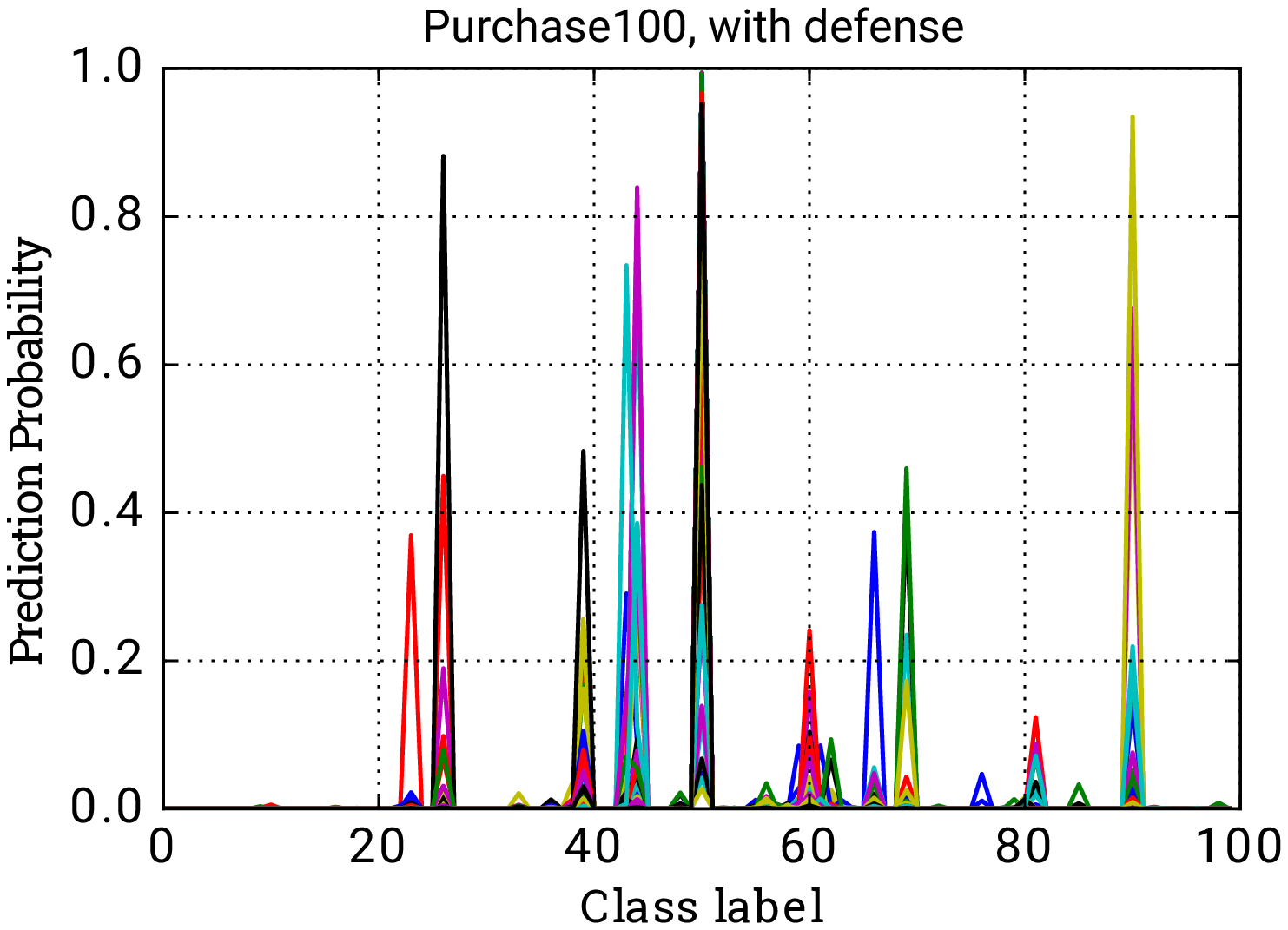}
    \caption{The distribution of the output (prediction vector) of the classifier on the \underline{training data} samples from class 50 in the Purchase100 dataset.  Each color represents one data sample. Without the defense, all samples are classified into class 50 with a probability close to 1.  Whereas, the privacy-preserving classifier spreads the prediction probability across many classes.  This added uncertainty is what provably mitigates the information leakage.  The figure at the bottom is computed on the \underline{test data} samples from class 50, which is indistinguishable from the middle figure.}\label{fig:pred_distribution50}
\end{figure}


\begin{figure*}[th!]
	\centering
	\includegraphics[width=0.9\columnwidth]{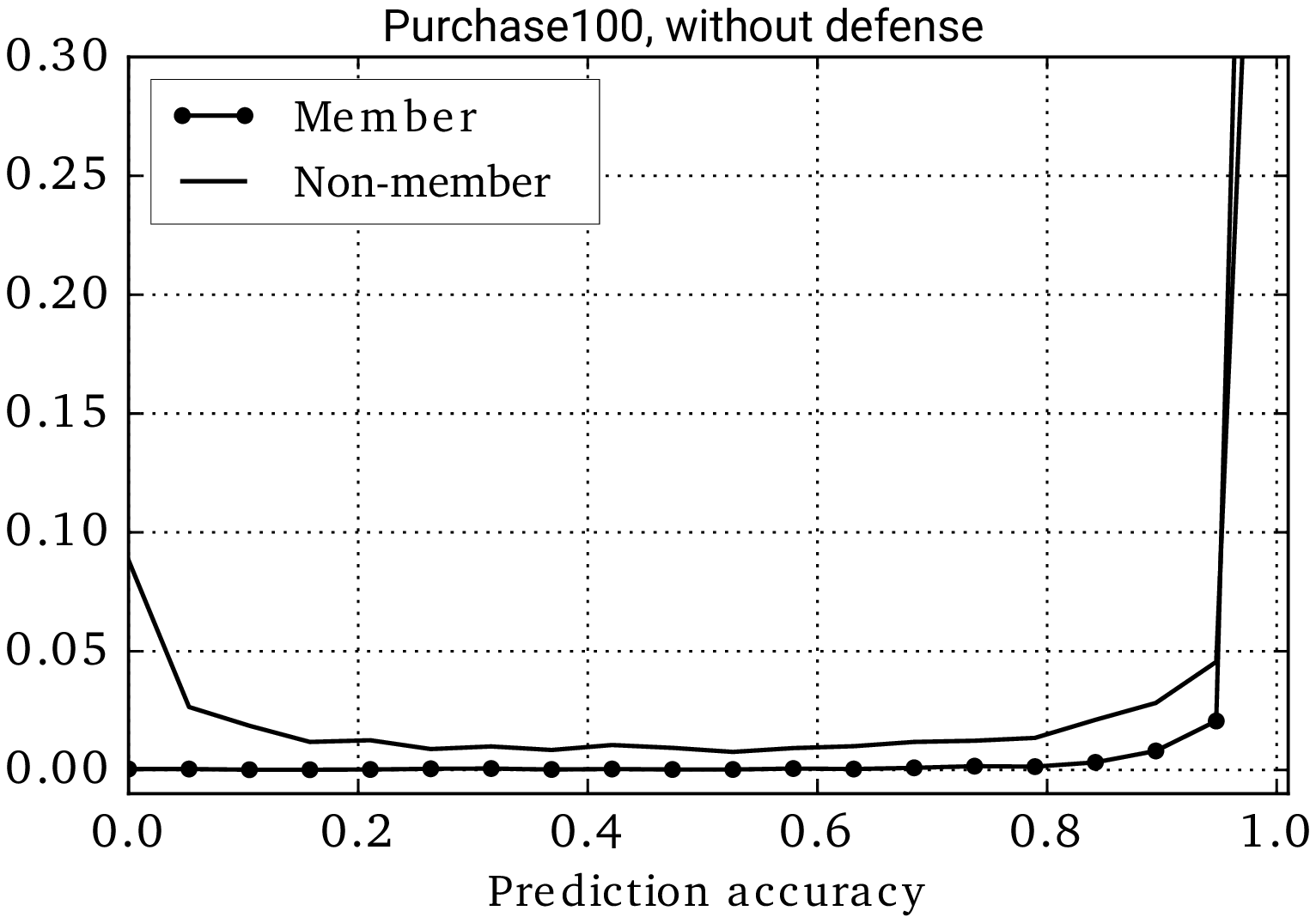}\qquad
	\includegraphics[width=0.9\columnwidth]{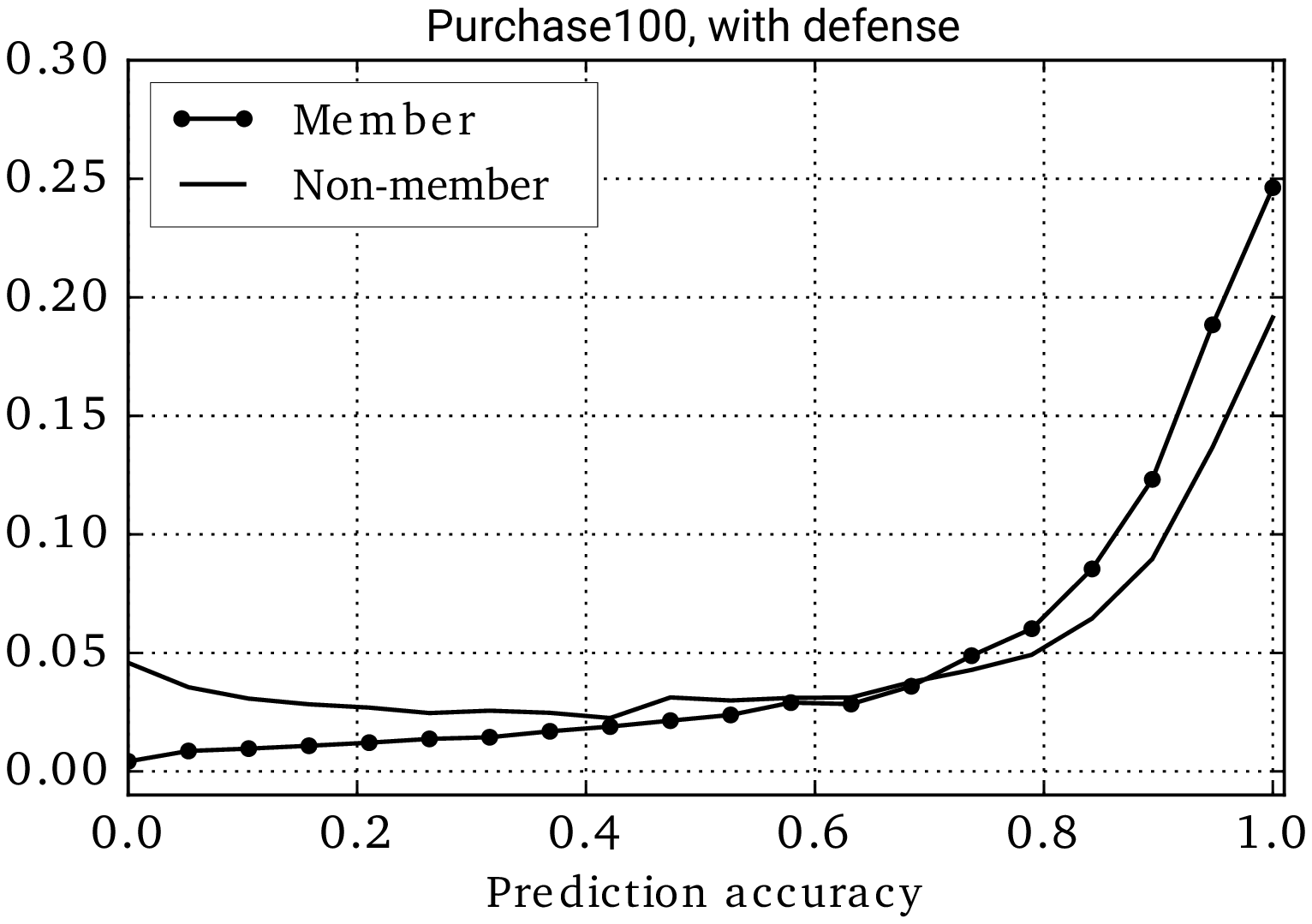}\\[15pt]
	\includegraphics[width=0.9\columnwidth]{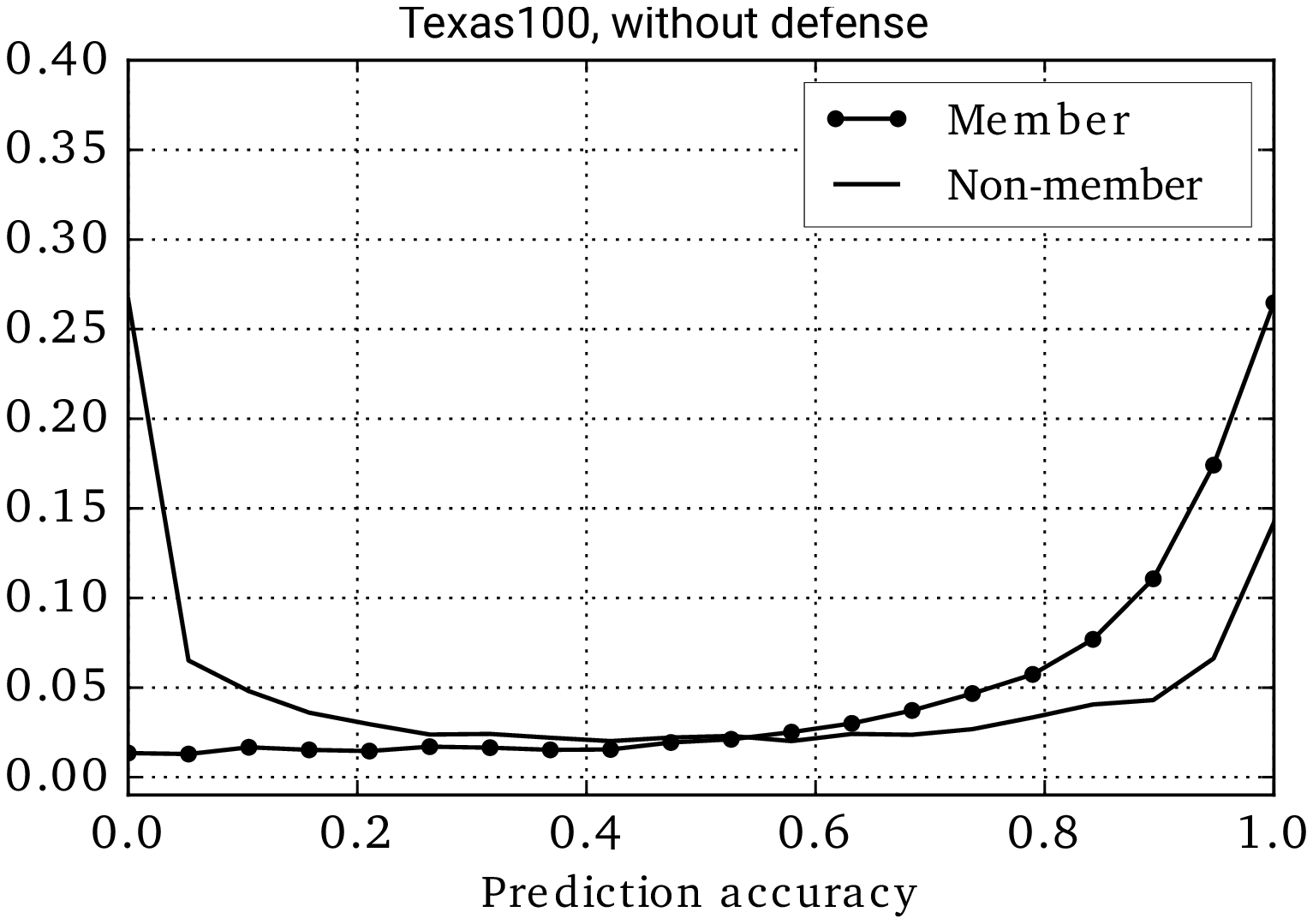}\qquad
	\includegraphics[width=0.9\columnwidth]{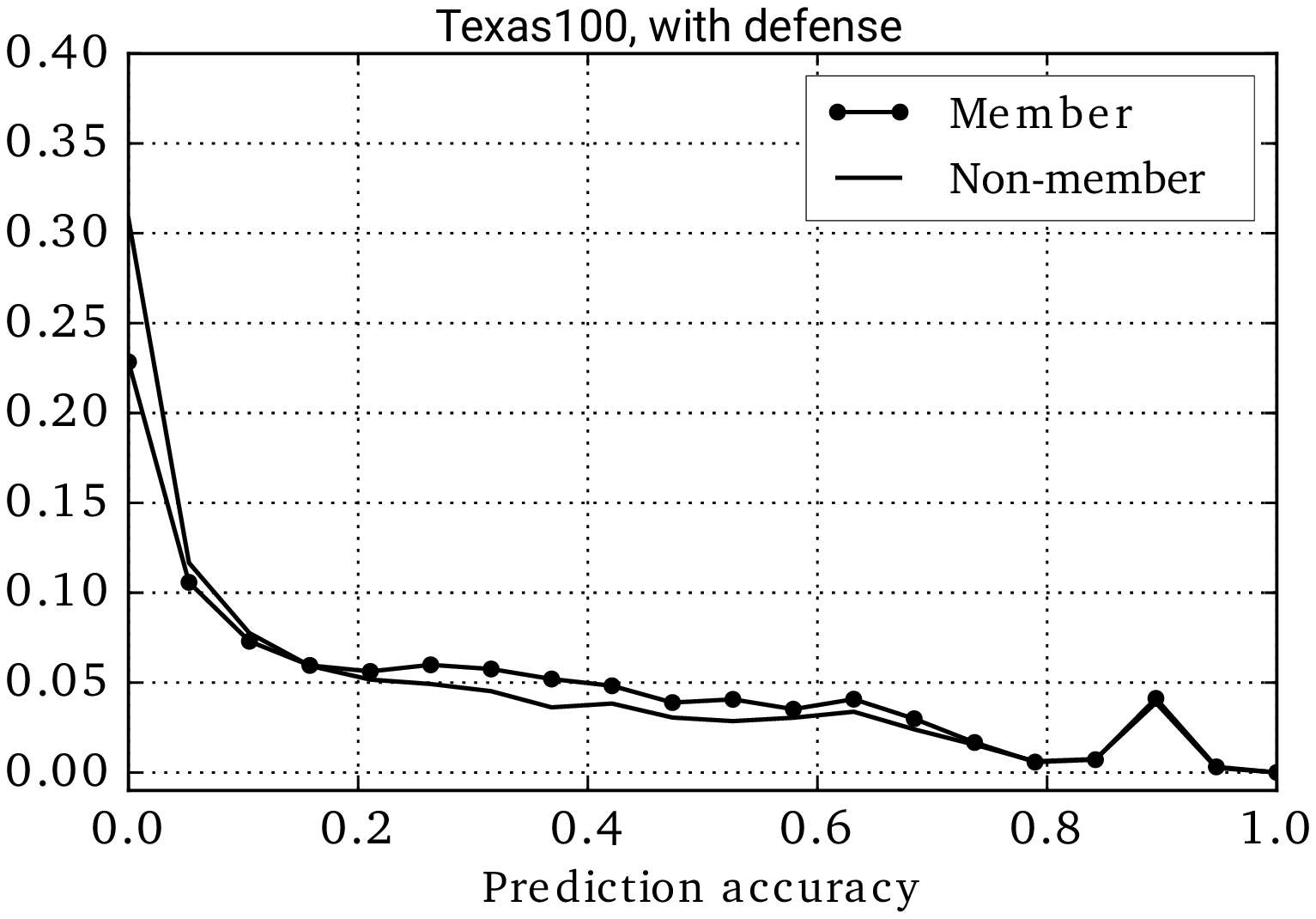}\\[15pt]
	\includegraphics[width=0.9\columnwidth]{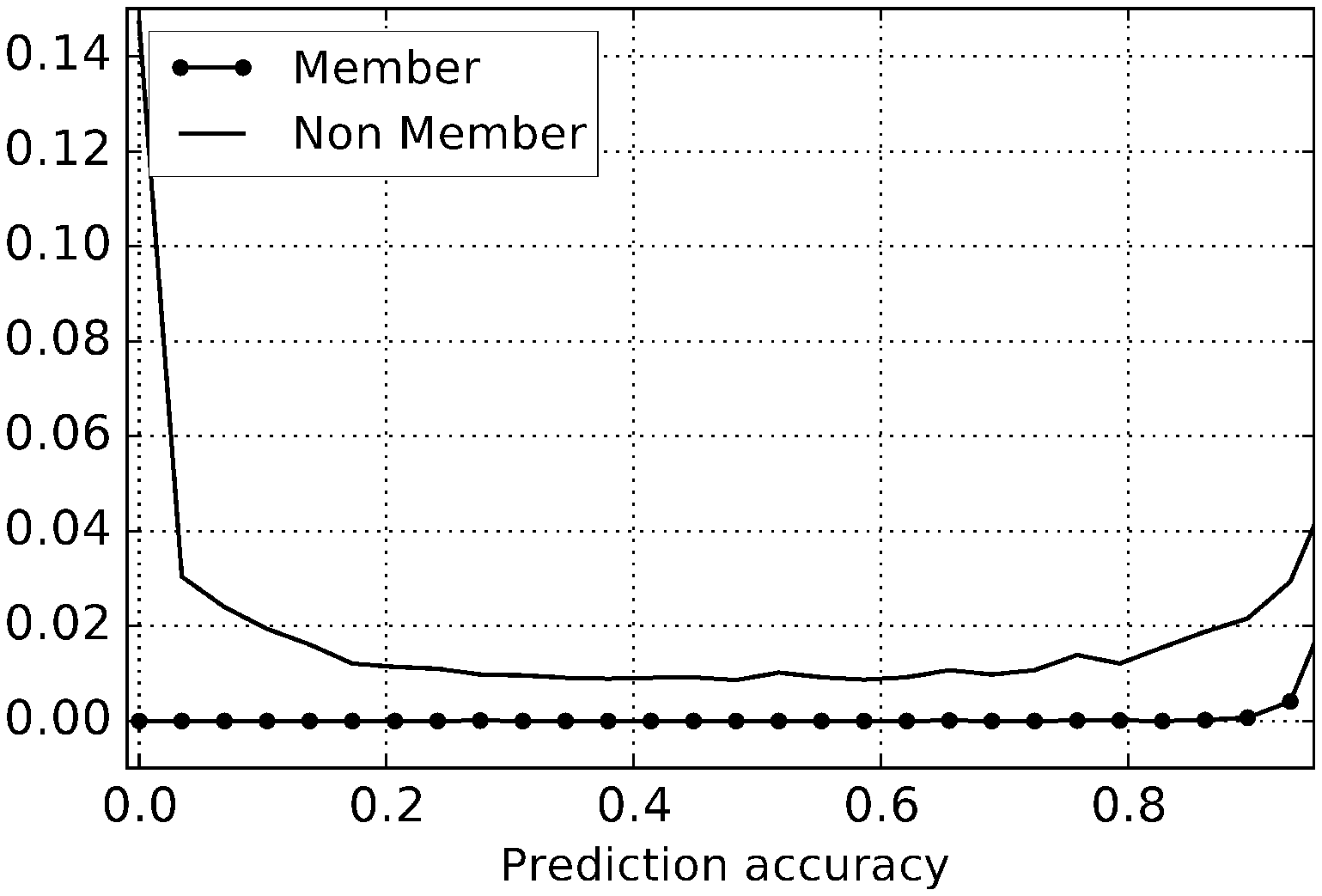}\qquad
	\includegraphics[width=0.9\columnwidth]{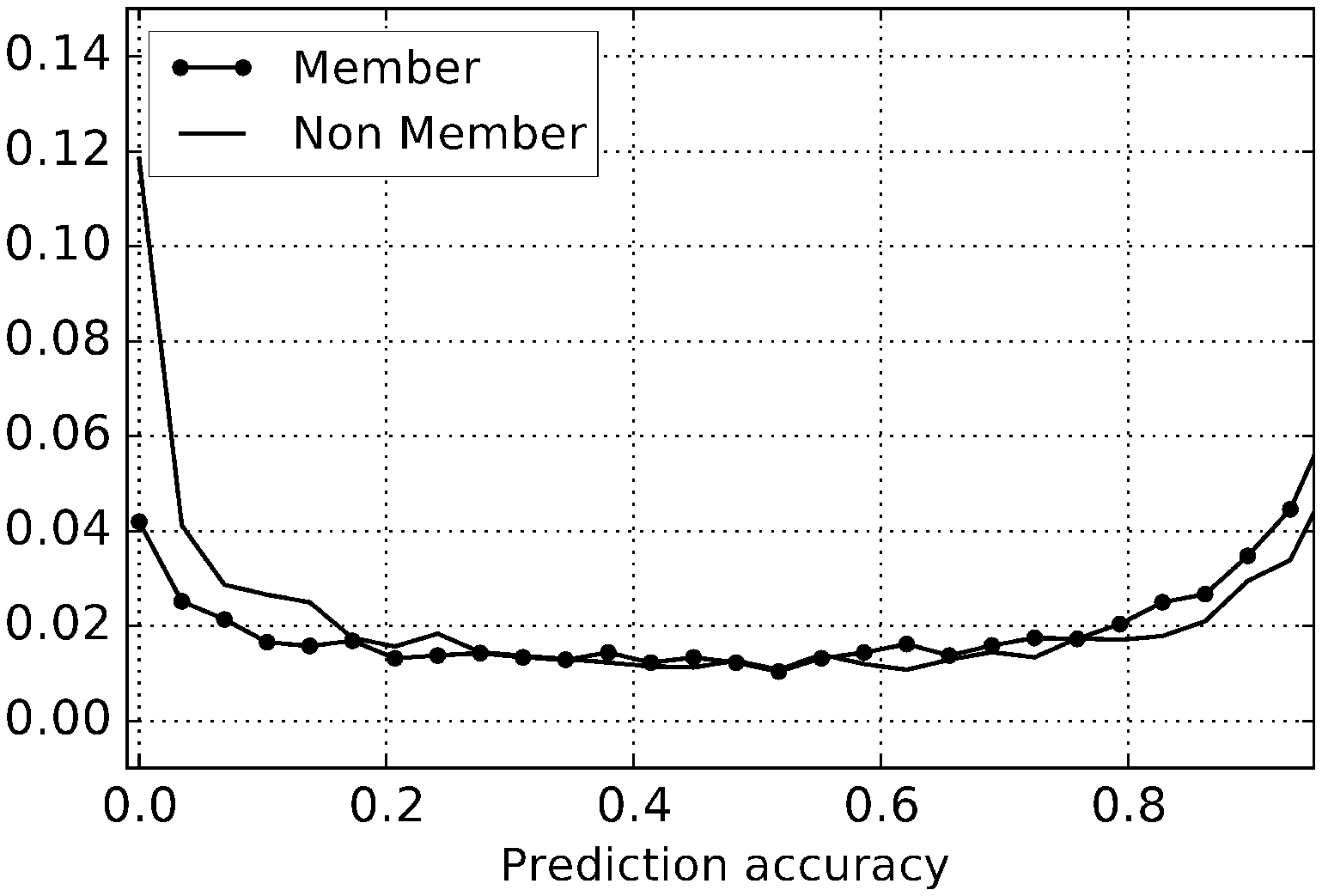}
	\caption{\underline{Distribution of the classifier's prediction accuracy on members of its training set versus non-member data samples}.  Accuracy is measured as the probability of predicting the right class for a sample input.  The plots on the left show the distribution curves for regular models (without defense), and the ones on the right show the distribution curves for privacy-preserving models (with defense).  The larger the gap between the curves in a plot is, the more the information leakage of the model about its training set is.  The privacy-preserving model reduces this gap by one to two orders of magnitude.\\[2pt]
	-- The {\em maximum} gap between the curves (with defense versus without defense) is as follows.\\  Purchase100 model: ($0.02$ vs. $0.34$),  Texas100 model: ($0.05$ vs. $0.25$),  and CIFAR100-Densenet model: ($0.06$ vs. $0.56$). \\[2pt]
	-- The {\em average} gap between the curves is as follows.\\  Purchase100 model: ($0.007$ vs. $0.013$), Texas100 model: ($0.004$ vs. $0.016$), and CIFAR100-Densenet model: ($0.005$ vs. $0.021$).\\[10pt]}\label{fig:pred_accuracy}
\end{figure*}


\begin{figure*}[th!]
    \centering
    \includegraphics[width=0.9\columnwidth]{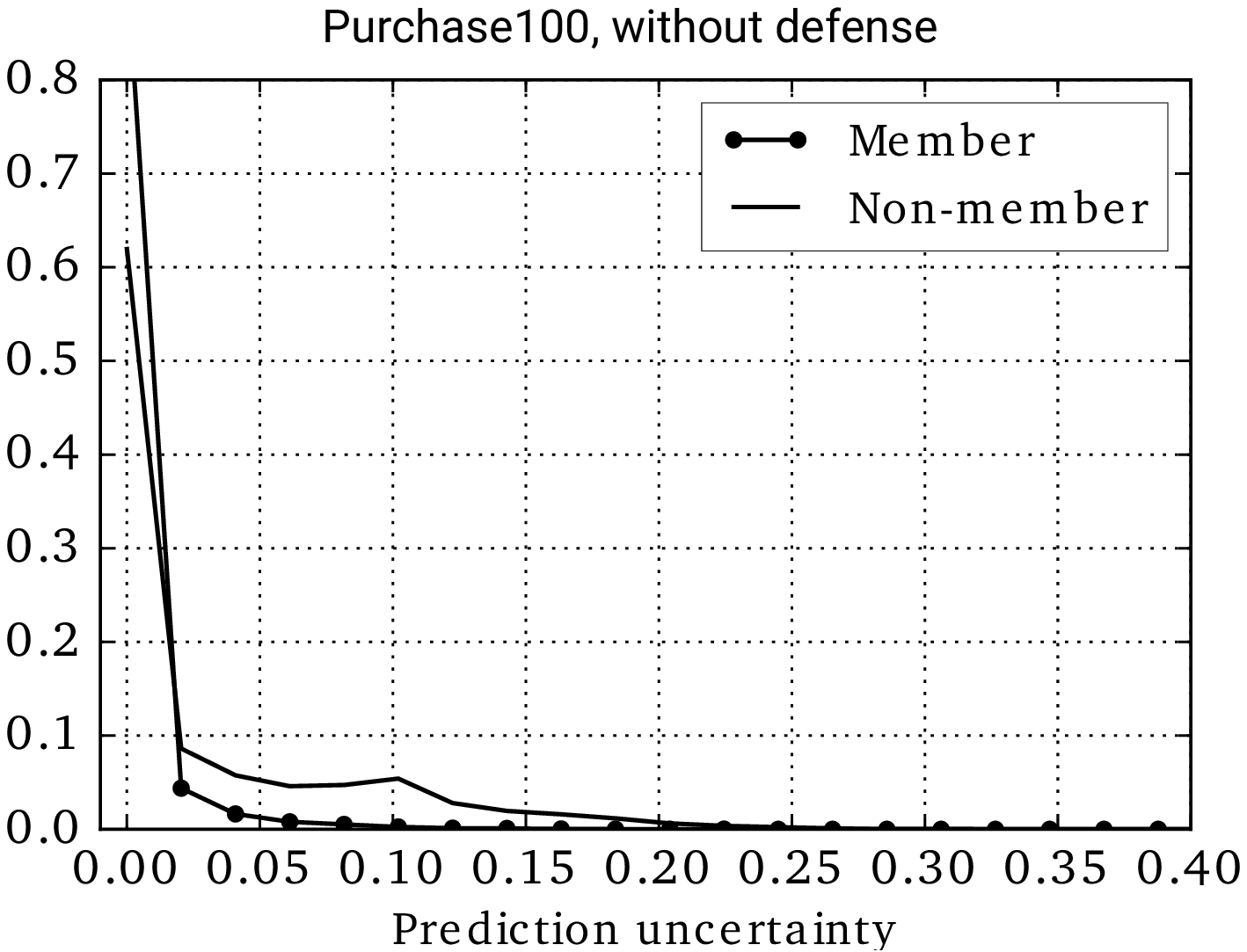}\qquad
    \includegraphics[width=0.9\columnwidth]{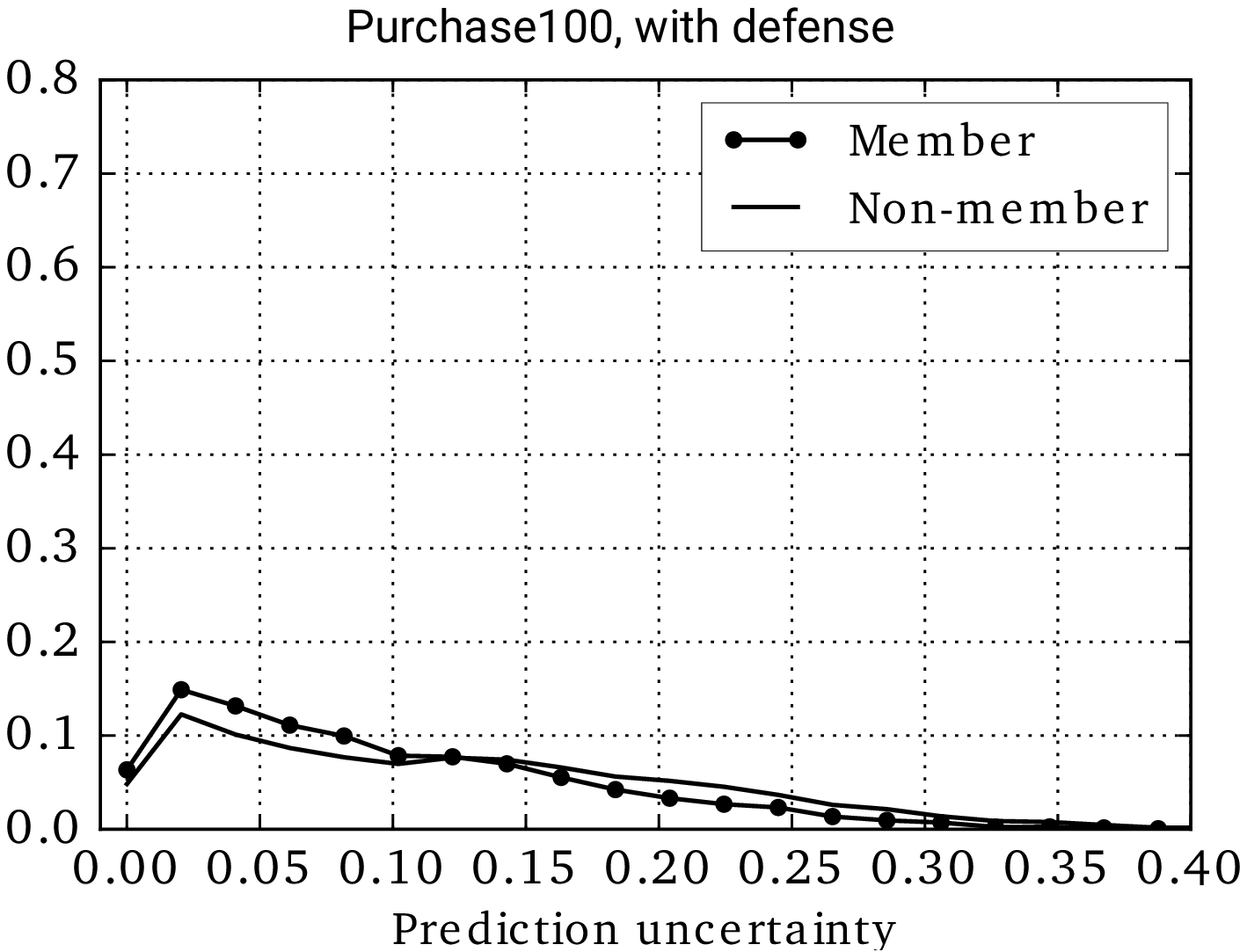}\\[15pt]
    \includegraphics[width=0.9\columnwidth]{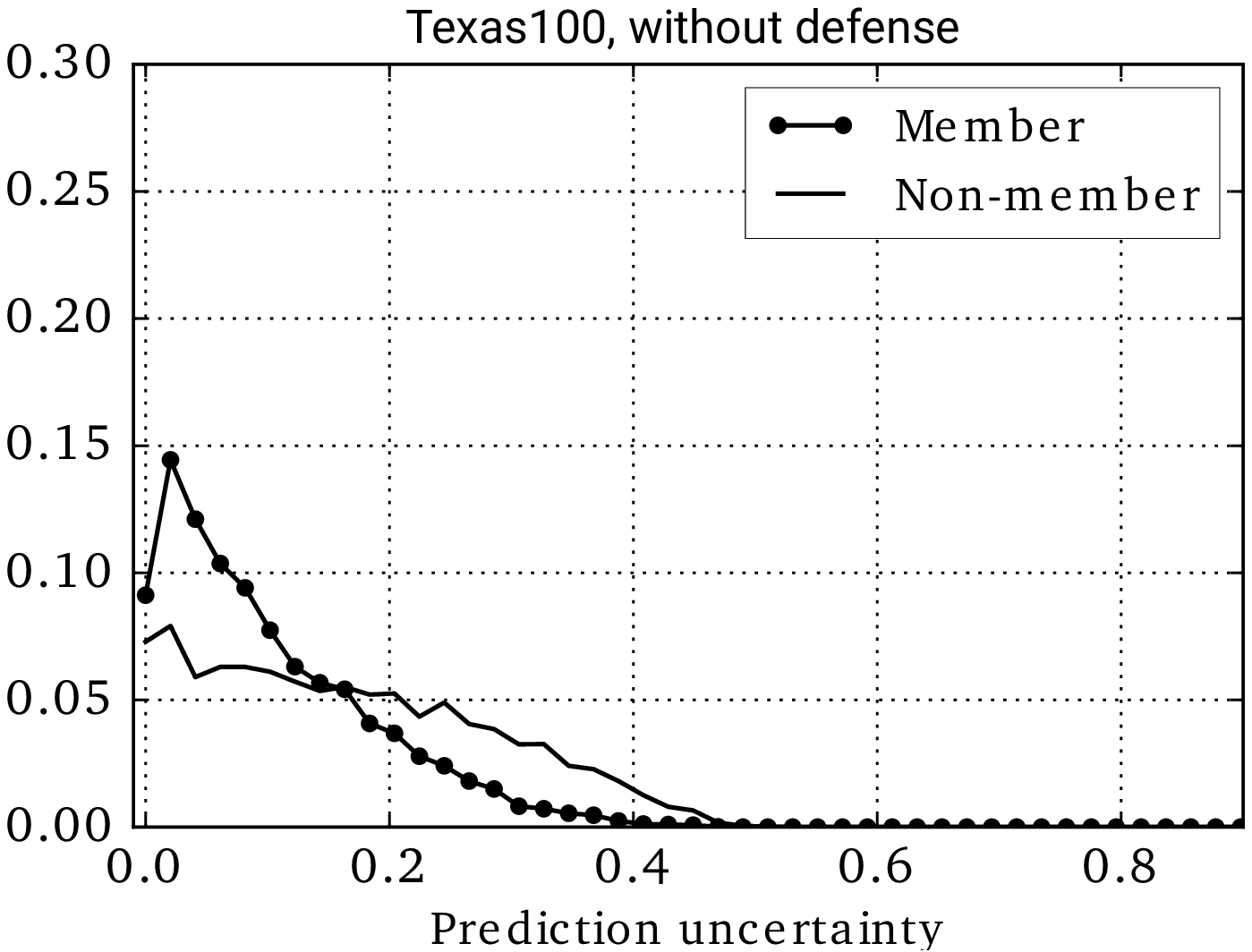}\qquad
    \includegraphics[width=0.9\columnwidth]{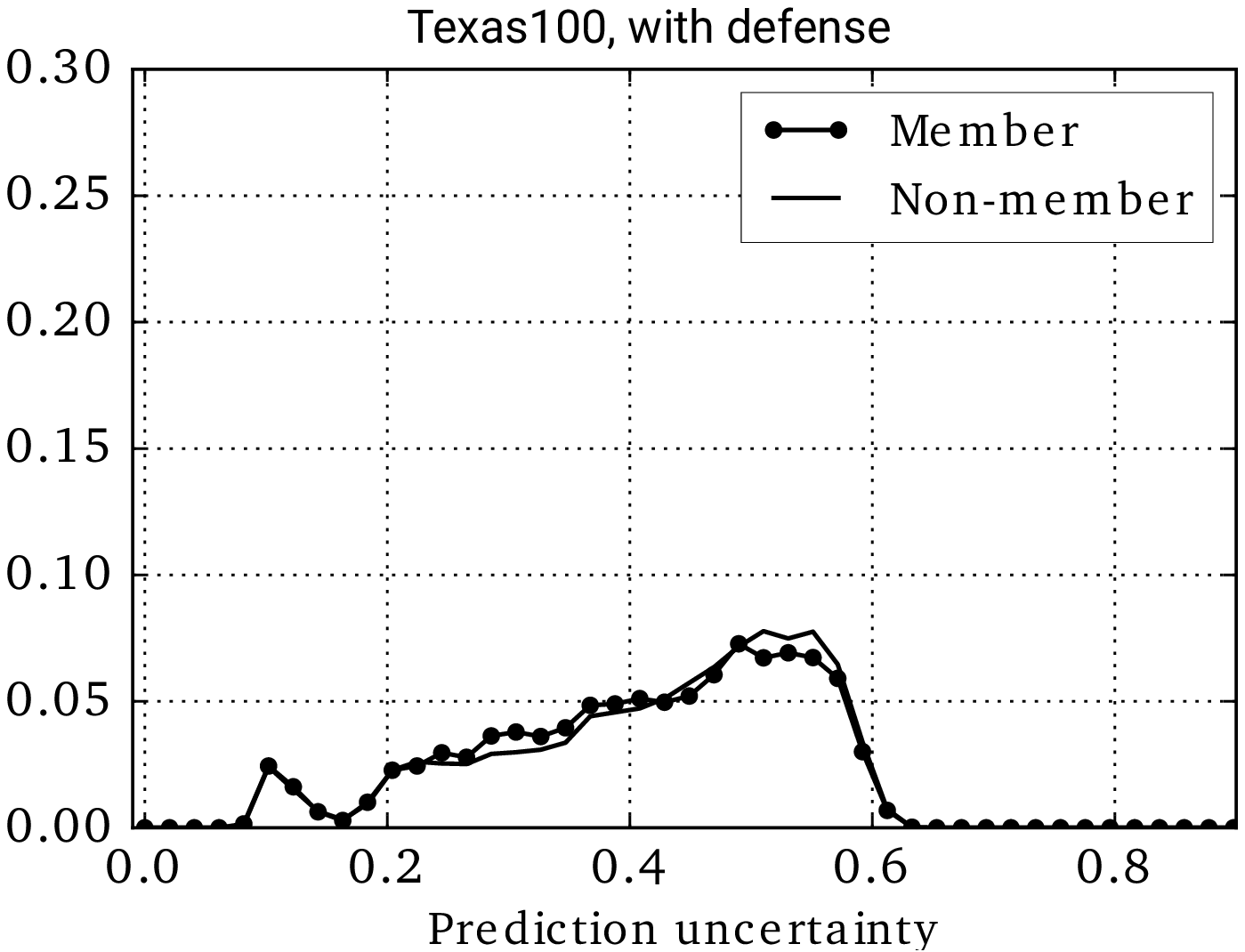}\\[15pt]
    \includegraphics[width=0.9\columnwidth]{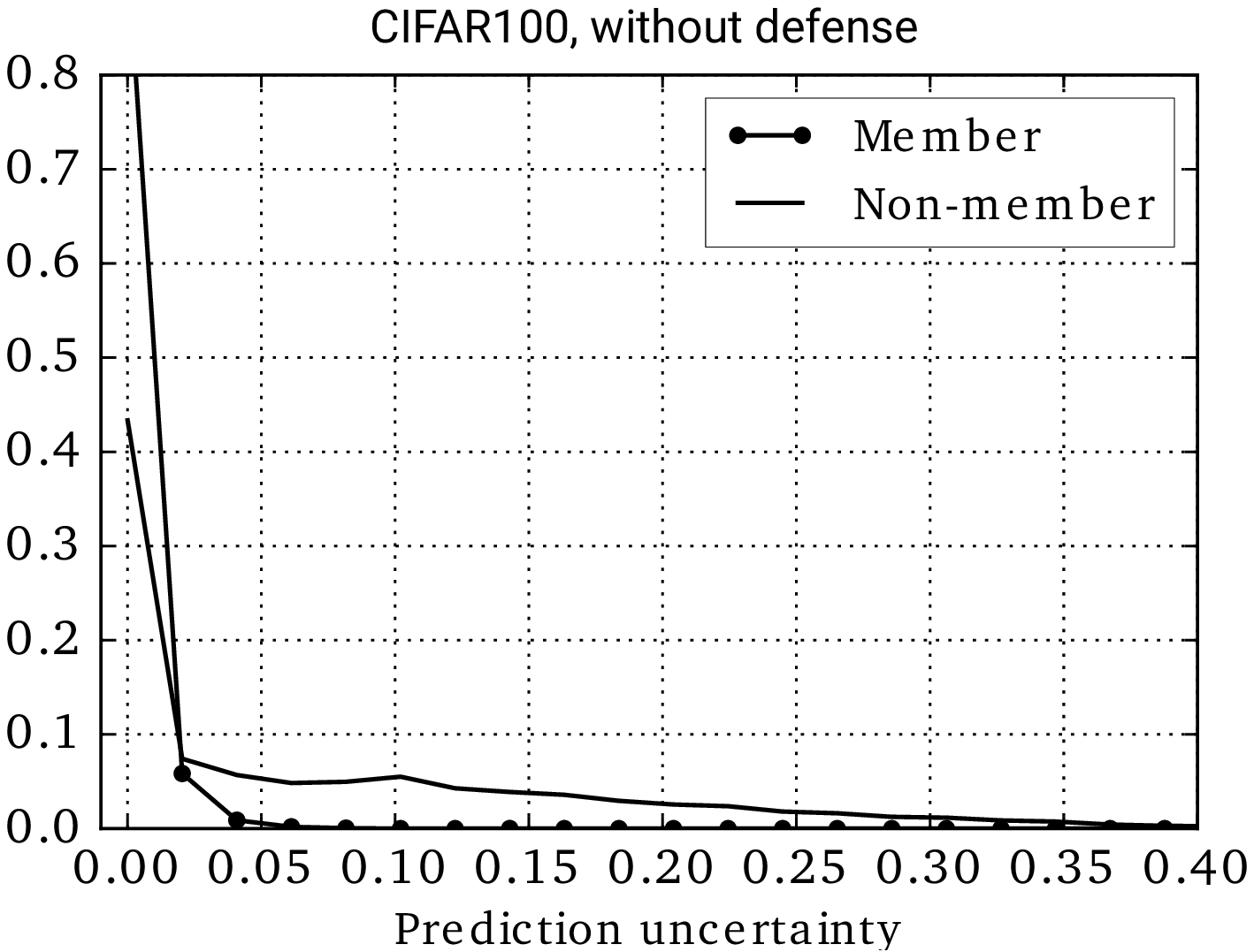}\qquad
    \includegraphics[width=0.9\columnwidth]{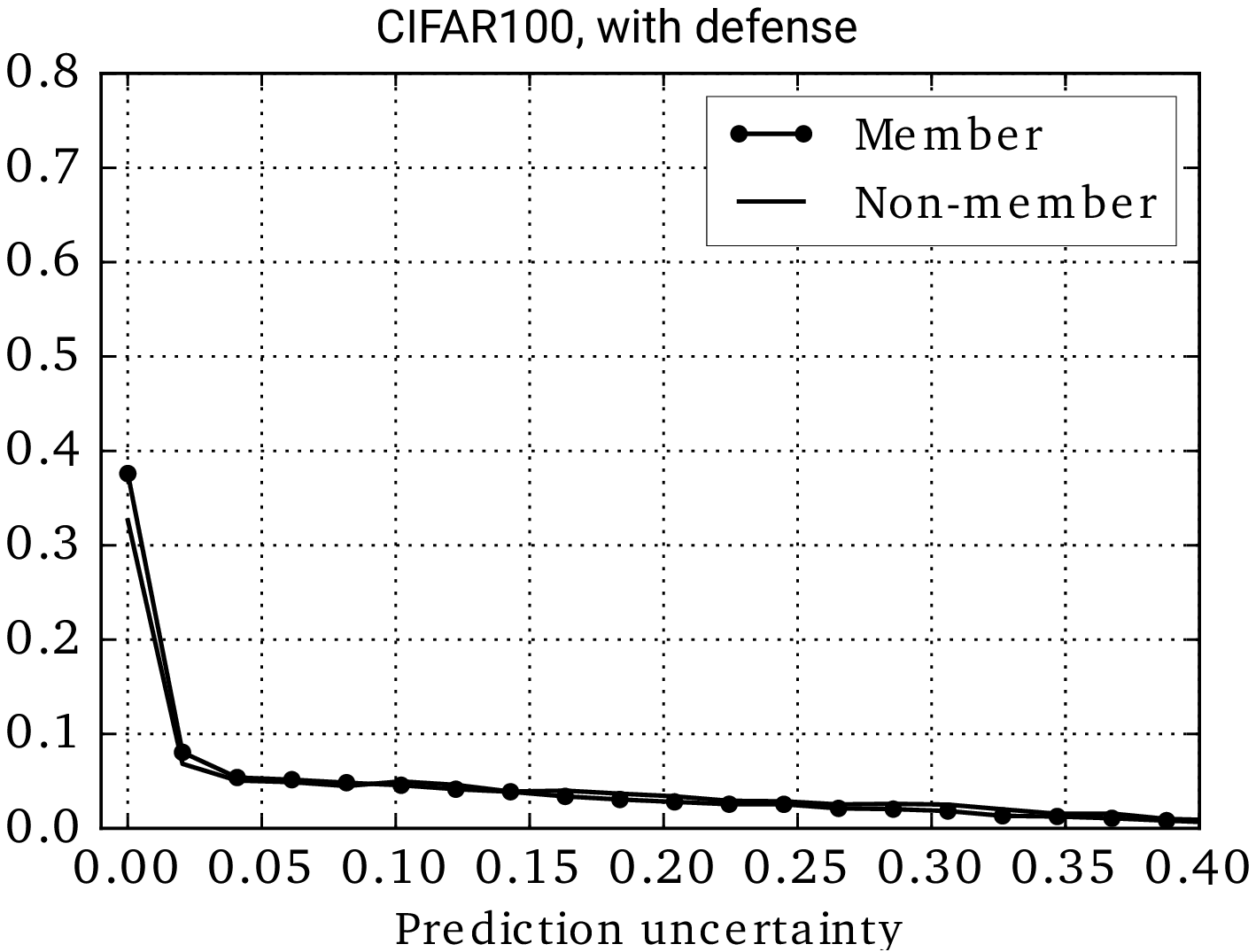}
    \caption{\underline{Distribution of the classifier's prediction uncertainty on members of its training set versus non-member data points}.  Uncertainty is measured as normalized Entropy of the model's output (i.e., prediction vector).  The plots on the left show the distribution curves for regular models (without defense), and the ones on the right show the distribution curves for privacy-preserving models (with defense).  The larger the gap between the curves in a plot is, the more the information leakage of the model about its training set is.  The privacy-preserving model reduces this gap by one to two orders of magnitude.\\[2pt]
    \qquad-- The {\em maximum} gap between the curves (with defense versus without defense) is as follows.\\  Purchase100 model: ($0.03$ vs. $0.30$),  Texas100 model: ($0.02$ vs. $0.15$),  and CIFAR100-Densenet model: ($0.04$ vs. $0.49$). \\[2pt]
    \qquad-- The {\em average} gap between the curves is as follows.\\  Purchase100 model: ($0.004$ vs. $0.012$), Texas100 model: ($0.002$ vs. $0.04$), and CIFAR100-Densenet model: ($0.002$ vs. $0.01$).\\[15pt]}\label{fig:pred_uncertainty}
\end{figure*}

\subsection{Empirical Results}

\paragraphb{\underline{Loss and Gain of the Adversarial Training}}

Figure~\ref{fig:trajectories} shows the empirical loss of the classification model~\eqref{eq:closs_emp} as well as the empirical gain of the inference model~\eqref{eq:gain_emp} throughout the training using Algorithm~\eqref{alg:training}.  By observing both the classifier's loss and the attack's gain over the training epochs, we can see that they converge to an equilibrium point.  Following the optimization problem \eqref{eq:game}, the attacker's gain is the maximum that can be achieved against the best defense mechanism. The classifier's loss is also the minimum that can be achieved while preserving privacy against the best attack mechanism.  As shown in Section~\ref{sec:game}, by minimizing the classifier's loss, we train a model that not only prevents the attack's gain to grow large, but also forces the adversary to play the random guess strategy.

In Figure~\ref{fig:trajectories}~(top), we compare the evolution of the classification loss of the privacy-preserving model with the loss of the same model when trained regularly (without defense).  As we can see, a regular model (without defense) arbitrarily reduces its loss, thus might overfit to its training data.  Later in this section, in Figure~\ref{fig:pred_distribution50}, we visualize the impact of this small loss on the output of the model, and we show how this can leak information about the model's training set.  {\bf\em The adversarial training for membership privacy strongly regularizes the model}.  Thus, our privacy-preserving mechanism not only protects membership privacy but also significantly prevents overfitting.

\paragraphb{\underline{Privacy and Generalization}}

To further study the tradeoff between privacy and predictive power of privacy-preserving models, in Figure~\ref{fig:genralizationerror} we show the cumulative distribution of the model's generalization error over different classes.  The plot shows the fraction of classes (y-axis) for which the model has a generalization error under a certain value (x-axis).  For each class, we compute the model's generalization error as the difference between the testing and training accuracy of the model for samples from that class~\cite{hardt2015train}.  We compare the generalization error of a regular model and our privacy-preserving model.  As the plots show, the generalization error of our privacy mechanism is significantly lower over all the classes.

Table~\ref{tab:results} presents all the results of training privacy-preserving machine learning models using our min-max game, for all our datasets.  It also compares them with the same models when trained regularly (without defense).  Note the gap between training and testing accuracy with and without the defensive training.  Our mechanism reduces the total generalization error by a factor of up to 4.  For example, the error is reduced from 29.7\% down to 7.5\% for the Texas100 model, it is reduced from 54.3\% down to 22.7\% for the CIFAR100-Alexnet model, and it is reduced from 29.4\% down to 12.7\% for the CIFAR100-Densenet model, while it remains almost the same for the Purchase100 model.  {\bf\em Our min-max mechanism achieves membership privacy with the minimum generalization error.}  Table~\ref{tab:lambda} shows how we can control the trade-off between prediction accuracy and privacy, by adjusting the adversarial regularization factor $\lambda$.

The regularization effect of our mechanism can be compared to what can be achieved using common regularizers such as the L2-norm regularizer, where $R(f_{\theta}) = \sum_i \theta_i^2$ (see our formalization of a classification loss function~\eqref{eq:coptimization}).  Table~\ref{tab:L2regularization} shows the tradeoff between the model's test accuracy and membership privacy using L2-norm.  Such regularizers do not guarantee privacy nor they minimize the cost of achieving it.  For a close-to-maximum degree of membership privacy, the testing accuracy of our privacy-preserving mechanism is {\em more than twice} the testing accuracy of a L2-norm regularized model.  This is exactly what we would expect from the optimization objectives of our privacy-preserving model.\\

\paragraphb{\underline{Membership Privacy and Inference Attack Accuracy}}

Table~\ref{tab:results} presents the training and testing accuracy of the model, as well as the attack accuracy.  To measure the {\em attack accuracy}, we evaluate the average probability that the inference attack model correctly predicts the membership:  
\begin{align*}
\frac{\sum\limits_{(x,y)\in\set{D}\setminus\DadvIN} h(x, y, f(x)) + \sum\limits_{(x'',y'')\in\set{D''}} (1 - h(x'', y'', f(x'')))}{|\set{D}\setminus\DadvIN| + |D''|}
\end{align*}
where $D''$ is a set of data points that are sampled from the same underlying distribution as the training set, but does not overlap with $\set{D}$ nor with $\DadvOUT$. 

The most important set of results in Table~\ref{tab:results} is the two pairs of colored columns which represent the testing accuracy of the classifier versus the attack accuracy.  There is a tradeoff between the predictive power of the model and its robustness to membership inference attack.  As expected from our theoretical results, the experimental results show that the attack accuracy is much smaller (and close to random guess) in the privacy-preserving model compared to a regular model.  {\bf\em Our privacy-preserving mechanism can guarantee maximum achievable membership privacy with only a negligible drop in the model's predictive power}.  To achieve a near maximum membership privacy, the testing accuracy is dropped by 3.5\% for the Purchase100 model, it is dropped by 4.4\% for the Texas100 model, it is dropped by 1.1\% for the CIFAR100-Alexnet model, and it is dropped by 3\% for the CIFAR100-Densenet model.\\

\paragraphb{\underline{Effect of the Reference Set}}

The objective of our min-max optimization is to make the predictions of the model on its training data indistinguishable from the model's predictions on any sample from the underlying data distribution.  We make use of a set of samples from this distribution, named reference set, to empirically optimize the min-max objective.  Table~\ref{tab:refsize} shows the effect of the size of the reference set $\set{D'}$ on the model's membership privacy.  The models are trained on the same training set $\set{D}$ of size 20,000, and hyper-parameter $\lambda = 3$.  As expected, as the size of the reference set increases, it becomes better at properly representing the underlying distribution, thus the attack accuracy converges to 50\%. \\

\paragraphb{\underline{Indistinguishability of Predictions}}

The membership inference attacks against black-box models exploit the statistical differences between the predictions of the model on its members versus non-members.  Figure~\ref{fig:pred_distribution50} shows the output of the model (i.e., the probability of being a sample from each class) on its training data, for a regular model (without defense) versus a privacy-preserving model.  The input data are all from class 50 in the Purchase100 dataset.  The top figure illustrates that a regular model (which is overfitted on its training set) produces a high probability for the correct class on its training data.  This significantly contributes to the vulnerability of the model to the membership inference attack.  The privacy-preserving model produces a visibly different distribution (the middle figure).  This makes the members' outputs indistinguishable from non-members' outputs (the bottom figure).  The min-max optimization makes these two output distributions converge to indistinguishable distributions.

We further investigate the indistinguishability of these two distributions by computing some statistics (accuracy and uncertainty) of the model's output for different datasets.  Figure~\ref{fig:pred_accuracy} and Figure~\ref{fig:pred_uncertainty} show the results as the histogram of the models' accuracy and uncertainty over the training set and testing set.  We compute the accuracy of model $f$ on data point $(x,y)$ as $f_y(x)$, which is the probability of predicting class $y$ for input $x$.  We compute uncertainty as the normalized entropy $\frac{-1}{\log(k)}\sum_i \hat{y}_i \log(\hat{y}_i)$ of the probability vector $\hat{y} = f(x)$, where $k$ is the number of classes.  The two figures show that {\bf\em our privacy mechanism significantly reduces both the maximum (worst case risk) and average gap between the prediction accuracy (and uncertainty) of the model on its training versus test set}, compared with a regular model.  Note that these figures do not prove privacy, but illustrate what the attacker can exploit in his inference attacks. They visibly show how the indistinguishability of the model's output distributions (on members and non-members) can improve by using our defense mechanism.


\section{Related Work}

Analyzing and protecting privacy in machine learning models against different types of attacks is a topic of ongoing research.  A direct privacy threat against machine learning is the untrusted access of the machine learning platform during training or prediction.  A number of defense mechanisms, which are based on trusted hardware and cryptographic private computing, have been proposed to enable blind training and use of machine learning models.  These methods leverage various techniques including homomorphic encryption, garbled circuits, and secure multi-party computation for private machine learning on encrypted data~\cite{mohassel2017secureml, gilad2016cryptonets, lindell2000privacy, bonawitz2017practical}, as well as private computation using trusted hardware (e.g., Intel SGX)~\cite{ohrimenko2016oblivious, hunt2018chiron}.  Although these techniques prevent an attacker from directly observing the sensitive data, yet they do not limit  information leakage through the computation itself. 

An adversary with some background knowledge and external data can try to infer information such as the training data, the input query, and the parameters of the model.  These inference attacks include input inference~\cite{fredrikson2015model}, membership inference~\cite{shokri2017membership}, attribute inference~\cite{carlini2018secret}, parameter inference~\cite{tramer2016stealing, wang2018stealing}, and side-channel attacks~\cite{wei2018know}.  There are examples of a wide-range of privacy attacks against computations over sensitive data.  Our focus is on the privacy risks of computation on databases, when the adversary observes the result of the computation.  In such settings, membership inference attacks and reconstruction attacks are considered as the two  major classes of attacks~\cite{dwork2017exposed}.

Membership inference attack is a decisional problem. It  aims at inferring the presence of a target data record in the (training) dataset~\cite{homer2008resolving, dwork2015robust, shokri2017membership, sankararaman2009genomic, backes2016membership, pyrgelis2017knock}.  The accuracy of the attack  shows the extent to which a model is dependent on its individual training data.  The reconstruction attack is a more generic type of attack, where the objective is to infer sensitive attributes of many individuals in the training set~\cite{dinur2003revealing, wang2009learning}.  One proposed defense technique against general inference attacks is computation (e.g., training of models) with differential privacy guarantee~\cite{dwork2006calibrating, dwork2014algorithmic}, which has  recently been used in the context of machine learning~\cite{chaudhuri2011differentially, bassily2014private, abadi2016deep, papernot2016semi,papernot2018scalable}.  Despite their provable robustness against inference attacks, differential privacy mechanisms are hard to achieve with negligible utility loss.  The utility cost comes from the fact that we aim at protecting privacy against all strong attacks by creating indistinguishability among similar states of all possible input datasets.  It is also related to the difficulty of computing a tight bound for the sensitivity of functions, which determines the magnitude of required noise for differential privacy.  The relation between some different definitions of membership privacy and differential privacy is analyzed in the literature~\cite{li2013membership, yeom2018privacy}.

Using game theory to formalize and optimize data privacy (and security) is another direction for protecting privacy~\cite{hsu2013differential, manshaei2013game, shokri2012protecting, alvim2017information,shokri2015privacygames}.  In such a framework, the privacy loss is minimized against the strongest corresponding attack.  The solution will be provably robust to any attack that threatens privacy according to such ``loss'' function.  The game-theoretic framework allows to explicitly incorporate the utility function into the min-max optimization, thus also minimizing the cost of the privacy defense mechanism.  The recent advances in machine learning, notably the developments of generative adversarial networks~\cite{goodfellow2014generative, arjovsky2017wasserstein, daskalakis2017training}, have introduced new algorithms for solving min-max games while training a complex (deep neural network) model.  Adversarial training has also been used for regularizing, hence generalizing, a model~\cite{kozinski2017adversarial, miyato2015distributional, miyato2017virtual, dai2017good, odena2016semi, dumoulin2016adversarially}.

\section{Conclusions}

We have introduced a new privacy mechanism for mitigating the information leakage of the predictions of machine learning models about the membership of the data records in their training sets.  We design an optimization problem whose objective is to jointly maximize privacy and prediction accuracy.  We design a training algorithm to solve a min-max game optimization that minimizes the classification loss of the model while maximizing the gain of the membership inference attack.  The solution will be a model whose predictions on its training data are indistinguishable from its predictions on any data sample from the same underlying distribution.  This mechanism guarantees membership privacy of the model's training set against the\textemdash strongest\textemdash inference attack, and imposes the minimum accuracy loss for achieving such level of privacy, given the available training/reference data and the capacity of the models.  In our extensive experiments on applying our method on benchmark machine learning tasks, we show that the cost of achieving privacy is negligible, and that our privacy-preserving models can generalize well.

\section*{Acknowledgements}

The authors would like to thank George Theodorakopoulos for helpful feedback.

\balance

\end{document}